\documentclass[10pt, twocolumn, letterpaper]{article}

\usepackage[pagenumbers]{iccv}

\usepackage[ruled, vlined]{algorithm2e}
\usepackage{multirow}
\usepackage[accsupp]{axessibility}
\newcommand{\sgets}{\mathrel{\scalebox{0.8}[0.8]{$\gets$}}}

\definecolor{iccvblue}{rgb}{0.21,0.49,0.74}
\usepackage[pagebackref, breaklinks, colorlinks, allcolors=iccvblue]{hyperref}

\def\paperID{10783}
\def\confName{ICCV}
\def\confYear{2025}

\title{Exploring Multimodal Diffusion Transformers for \\ Enhanced Prompt-based
Image Editing}

\author{Joonghyuk Shin \quad Alchan Hwang \quad Yujin Kim \quad Daneul Kim \quad
Jaesik Park\\ Seoul National University\\
{\tt\small \{joonghyuk, alchan00, yujin.k, carpedkm, jaesik.park\}@snu.ac.kr} }

\begin{document}
	\twocolumn[{ \maketitle \begin{center}\centering \captionsetup{type=figure} \includegraphics[width=\linewidth]{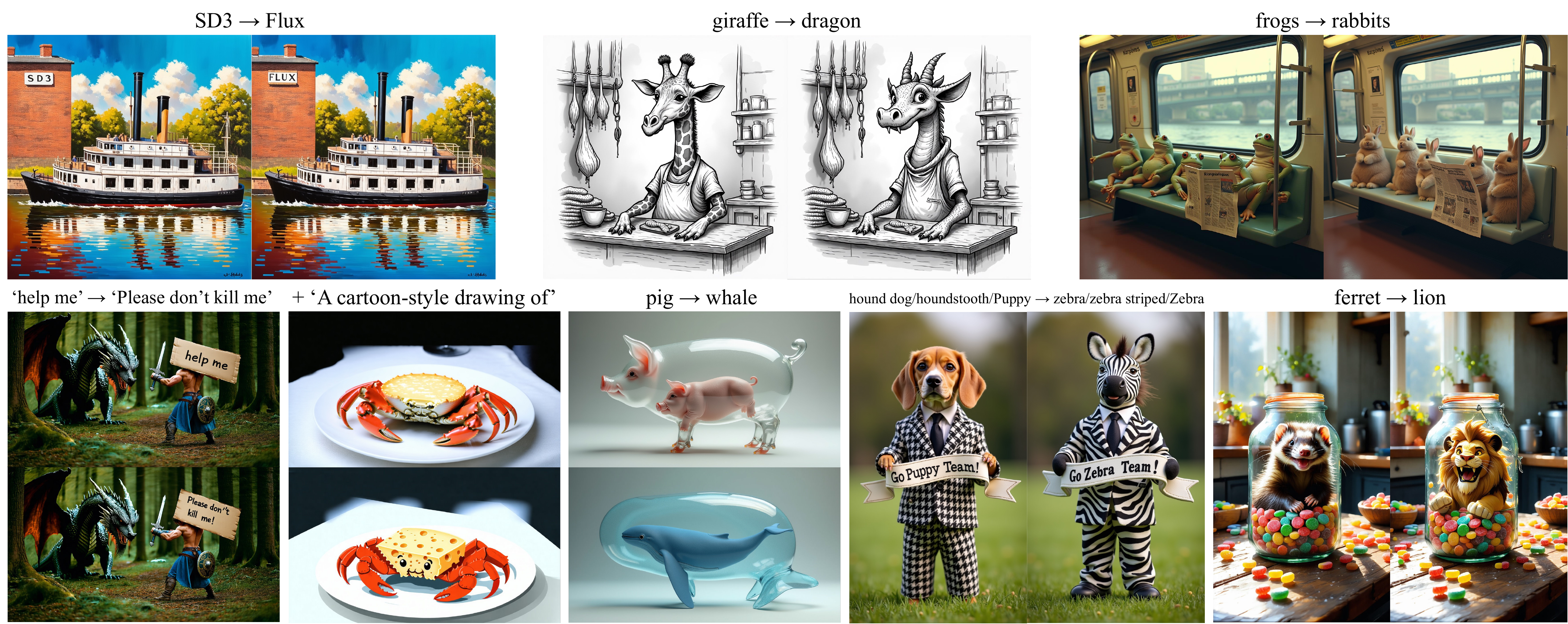} \caption{Our prompt-based editing examples on SD3 series and Flux.1, demonstrating high-quality results across various editing scenarios.} \label{Fig:teaser}\end{center} }]

	\begin{abstract}
		Transformer-based diffusion models have recently superseded traditional U-Net
		architectures, with multimodal diffusion transformers (MM-DiT) emerging as
		the dominant approach in state-of-the-art models like Stable Diffusion 3 and
		Flux.1. Previous approaches have relied on unidirectional cross-attention
		mechanisms, with information flowing from text embeddings to image latents.
		In contrast, MM-DiT introduces a unified attention mechanism that concatenates
		input projections from both modalities and performs a single full attention operation,
		allowing bidirectional information flow between text and image branches. This
		architectural shift presents significant challenges for existing editing techniques.
		In this paper, we systematically analyze MM-DiT's attention mechanism by decomposing
		attention matrices into four distinct blocks, revealing their inherent
		characteristics. Through these analyses, we propose a robust, prompt-based image
		editing method for MM-DiT that supports global to local edits across various
		MM-DiT variants, including few-step models. We believe our findings bridge the
		gap between existing U-Net-based methods and emerging architectures,
		offering deeper insights into MM-DiT's behavioral patterns.
	\end{abstract}

	\section{Introduction}
	\label{Sec:introduction} Recent years have witnessed the emergence of
	diffusion models~\cite{ho2020denoising, song2021scorebased} as the de-facto
	approach for visual generation, demonstrating remarkable capabilities in creating
	high-quality data from noise~\cite{rombach2022high, ramesh2022hierarchical,
	saharia2022photorealistic, podellsdxl, esser2024scaling, nichol2021improved,
	karras2022elucidating, luo2022understanding}. While early implementations
	predominantly employed U-shaped networks~\cite{ronneberger2015u}, transformer-based
	architectures~\cite{peebles2023scalable, chenpixart} have recently emerged as the
	prevailing approach, with multimodal diffusion transformers (MM-DiT) showing
	particular promise in state-of-the-art models like Stable Diffusion 3 series~\cite{esser2024scaling}
	and Flux.1~\cite{flux2024}.

	Motivated by MM-DiT’s architectural differences and the challenges in directly
	applying existing U-Net-based editing methods, our work addresses a
	fundamental question: ``How can we leverage our understanding of U-Net attention
	mechanisms to interpret MM-DiT’s behavior better and effectively transfer existing
	techniques to this new architecture?'' Given the significance of existing work
	built upon U-Net attention mechanisms~\cite{hertz2022prompt, cao2023masactrl,
	liu2024towards, chefer2023attend}, answering this question is crucial for enabling
	existing methods to adapt to, and benefit from, emerging MM-DiT architectures.

	To this end, we systematically investigate MM-DiT's attention mechanisms
	through block matrix analysis, revealing key characteristics of its attention maps.
	As one of our key analytical insights, we observe an interesting phenomenon aligned
	with vision transformer~\cite{dosovitskiy2020image} scaling laws~\cite{darcetvision}:
	as the model size increases, attention patterns become properly positioned but
	increasingly noisy, necessitating careful selection of attention layers for
	editing tasks. Building upon these analyses, we propose an efficient editing
	method at the architecture level that focuses on modifying image tokens in
	input projections. The proposed editing method can be jointly combined with inversion
	methods~\cite{rout2024semantic, wang2024taming} for editing real images as well.

	Our contributions are primarily architectural: (1) We provide the first
	systematic analysis of MM-DiT attention mechanisms, highlighting block-wise patterns
	and scaling behaviors. (2) Building upon these findings, we propose an efficient
	and robust editing method across diverse editing scenarios - from global stylistic
	changes to local edits such as text modification - and various MM-DiT variants,
	including few-step models (SD3-M, SD3.5-M, SD3.5-L, SD3.5-L-Turbo, and Flux.1-dev/schnell).

	\section{Related Work}
	\subsection{Text-to-image diffusion models}
	\label{Sec:related-work-text-to-image} Early text-to-image diffusion models
	mostly used U-shaped networks with residual blocks~\cite{ronneberger2015u}
	incorporating pixel-wise self-attention and text-image cross-attention layers,
	combined with downsampling and upsampling operations (\cref{Fig:unet-dit-architectures}).
	These models are typically trained with DDPM-style scheduling and noise
	prediction objective~\cite{rombach2022high, podellsdxl}.

	Recent advances introduced two significant shifts. First, architecturally, there
	has been a transition from U-Nets to transformer-based architectures like DiT~\cite{peebles2023scalable},
	demonstrating superior scaling properties. The PixArt-alpha series~\cite{chenpixart,
	chen2024pixart, chen2024pixartdelta} successfully adapted DiT for text conditional
	generation, maintaining a conventional cross-attention mechanism for text
	conditioning. Stable Diffusion 3~\cite{esser2024scaling} subsequently introduced
	a pivotal shift with multimodal diffusion transformers (MM-DiT), using
	separate transformers for text and image modalities while concatenating their
	sequences for unified attention operations (\cref{Fig:unet-dit-architectures}).
	Second, formulation-wise, newer approaches have adopted v-prediction~\cite{salimansprogressive}
	with a rectified flow framework~\cite{lipmanflow,liuflow} that connects data and
	noise distributions along straight paths, replacing the complex trajectories
	of traditional diffusion models. Together with increased model scales, these
	developments have significantly improved model performance. Parallel efforts have
	also accelerated iterative diffusion sampling via faster ODE solvers~\cite{lu2022dpm,
	lu2022dpmpp}, consistency distillation~\cite{song2023consistency, luo2023lcm},
	adversarial learning~\cite{sauer2023adversarial, yin2024one, yin2024improved, kang2024distilling},
	and trajectory straightening~\cite{lipmanflow, liuflow, liu2023instaflow}.
	\begin{table}
		\centering
		\caption{Architectural details and inference times for $1024 \times 1024$
		image generation using 28 timesteps, using single A6000 GPU. Times are shown
		as optimized (left, using PyTorch SDPA kernels) vs. naive matrix
		multiplication (right). Text encoders: CLIP\textsuperscript{1} refers to
		CLIP ViT-L/14~\cite{radford2021learning}, CLIP\textsuperscript{2} refers to OpenCLIP
		ViT-G/14~\cite{openclipilharco}, and T5 refers to T5-XXL~\cite{raffel2020exploring}.}
		\label{Tab:architecture_details}
		\vspace{-1mm}
		\footnotesize
		\begin{tabular}{@{}lcccc@{}}
			\toprule     & Arch. / Params  & Text Enc.                      & Attn Type  & Time (s)  \\
			\midrule SD1 & U-Net / 860M    & CLIP\textsuperscript{1}        & Self/Cross & 5.0/12.7  \\
			SDXL         & U-Net / 2.6B    & CLIP\textsuperscript{1,2}      & Self/Cross & 5.6/7.9   \\
			SD3-M        & MM-DiT / 2B     & CLIP\textsuperscript{1,2} / T5 & Full       & 7.1/20.5  \\
			SD3.5-M      & MM-DiT-X / 2.2B & CLIP\textsuperscript{1,2} / T5 & Full/Self  & 9.6/26.8  \\
			SD3.5-L      & MM-DiT / 8B     & CLIP\textsuperscript{1,2} / T5 & Full       & 24.7/60.5 \\
			Flux.1-dev   & MM-DiT* / 11.9B & CLIP\textsuperscript{1} / T5   & Full       & 27.8/36.8 \\
			\bottomrule
		\end{tabular}
		\vspace{-3.5mm}
	\end{table}

	\subsection{Text-guided image editing}
	\label{Sec:related-work-text-guided-edit}

	Text-guided image editing naturally extends text-to-image diffusion models.
	Prior work~\cite{liu2024towards, wang2023DPL, guo2024focus, hong2023improving}
	shows that self-attention governs spatial structure while cross-attention
	handles attribute-level semantics aligned with text. Prompt-to-Prompt (P2P)~\cite{hertz2022prompt},
	a seminal approach, leverages this by directly transferring cross-attention maps
	for precise edits. Subsequent methods further manipulate self-attention
	mechanisms for better content-preserving edits~\cite{cao2023masactrl, nam2024dreammatcher,
	jeong2024visual, zhou2024storydiffusion}. A key advantage of these attention-based
	methods is their ability to achieve precise local edits solely from textual prompts,
	eliminating the need for explicit spatial guidance or additional user input.

	Another direction employs inversion techniques~\cite{rout2024semantic, huberman2024edit,
	garibi2024renoise, deng2024fireflow, wang2024taming, jiao2025uniedit}, mapping
	real images to editable latent spaces. While inversion alone can lose fine
	details, combining it with attention-based methods significantly enhances localized
	editing~\cite{mokady2023null}. Alternatively, methods like InstructPix2Pix~\cite{brooks2023instructpix2pix}
	train dedicated models on synthetic data, though this approach incurs high
	computational costs and dataset preparation overheads~\cite{wu2024turboedit}.

	\subsection{MM-DiT variants}
	\label{Sec:related-work-mm-dit-variants}
	\begin{figure}
		\centering
		\includegraphics[width=\linewidth]{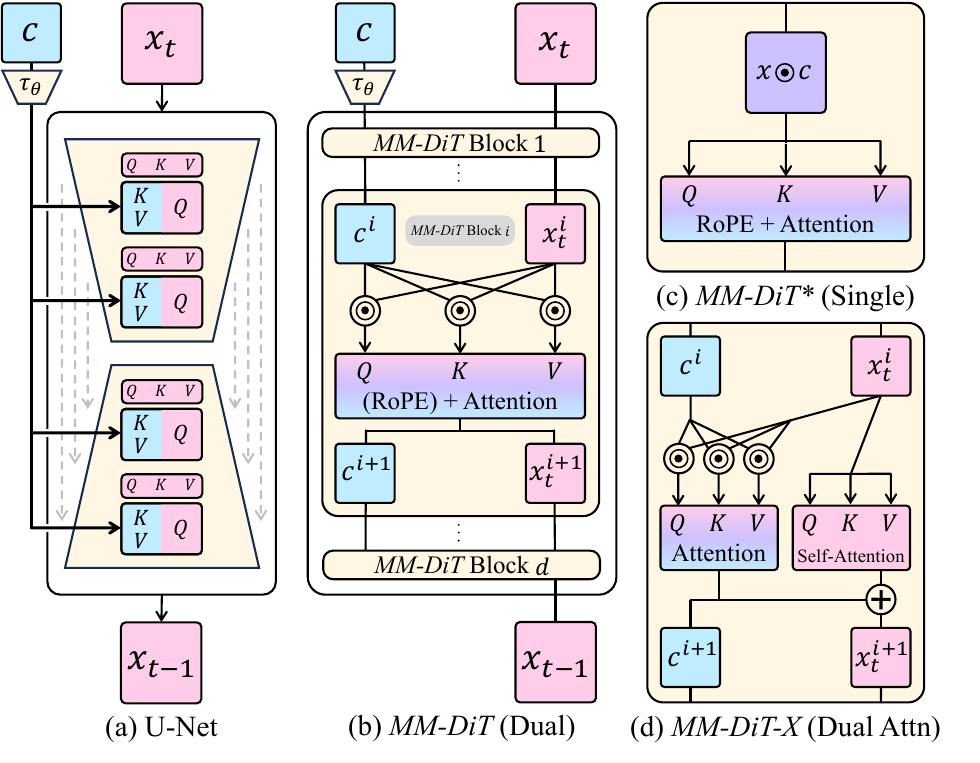}
		\caption{Attention mechanisms across diffusion architectures. (a) U-Net with
		separate self-/cross- attention, (b) MM-DiT blocks with dual branches, (c) MM-DiT*
		single branch block used in later layers of Flux.1, and (d) MM-DiT-X dual
		attention block with explicit self-attention, used in early layers of SD3.5-M.}
		\label{Fig:unet-dit-architectures}
		\vspace{-3mm}
	\end{figure}
	Since its introduction, MM-DiT architecture has seen several improvements. The
	initial dual-branch design in SD3 was motivated by the conceptual differences
	between text and image embeddings. This architecture was maintained in both SD3
	and SD3.5-Large models, which contain 24 and 38 dual-branch blocks,
	respectively. Flux.1, scaled to 12B parameters, introduced a hybrid approach
	with 57 total blocks - 19 dual branch blocks followed by 38 single branch
	blocks (\cref{Fig:unet-dit-architectures}). The single branch variant processes
	concatenated text and image embeddings as a single modality using unified weights
	(\textit{e.g.}, linear layers, attention parameters), reducing block parameters
	from 340M to 141M. The MM-DiT-X variant, later introduced in SD3.5-medium, added
	explicit self-attention operations exclusively for image tokens in the first 13
	transformer blocks for better multi-resolution results and overall image coherence.

	\section{Attention Analysis in MM-DiT}
	\subsection{Block matrix formulation of MM-DiT attention}
	\label{Sec:attention-analysis-interpret} We begin the analysis with MM-DiT's attention
	structure under SD3-M's standard dimensions: 64$\times$64 image latents (flattened
	to 4096) and 333 text tokens (77 CLIP + 256 T5), with inner dimension 64. Omitting
	the batch and head dimensions for clarity, the input projections for image and
	text domains are denoted as
	$(\mathbf{q}_{i}, \mathbf{k}_{i}, \mathbf{v}_{i}\in \mathbb{R}^{4096 \times 64}
	)$
	and
	$(\mathbf{q}_{t}, \mathbf{k}_{t}, \mathbf{v}_{t}\in \mathbb{R}^{333 \times 64})$.
	Then, query, key, and value projections are concatenated as shown below ($\mathbf{q}
	, \mathbf{k}, \mathbf{v}\in \mathbb{R}^{4429 \times 64}$). It is worth noting that
	Flux.1 adopts an inverse concatenation order (text preceding image), which results
	in a different attention map.
	\begin{align}
		\mathbf{q}= \begin{bmatrix}\mathbf{q}_{i}\\ \mathbf{q}_{t}\end{bmatrix}, \mathbf{k}= \begin{bmatrix}\mathbf{k}_{i}\\ \mathbf{k}_{t}\end{bmatrix}, \mathbf{v}= \begin{bmatrix}\mathbf{v}_{i}\\ \mathbf{v}_{t}\end{bmatrix}. \label{Eq:qkv}
	\end{align}
	Using the standard scaled dot product attention formulation
	$softmax\left(\frac{\mathbf{qk}^{T}}{\sqrt{d_{k}}}\right)\mathbf{v}$, we can write
	the attention map $\mathbf{qk}^{T}$ and its output representation as follows.
	Note that row-wise softmax normalization is applied to $\mathbf{qk}^{T}$ in~\cref{Eq:attn_map},
	before multiplication with $\mathbf{v}$ to obtain~\cref{Eq:attn_result}.
	\begin{align}
		 & \mathbf{q}\mathbf{k}^{T}= \begin{bmatrix}\mathbf{q}_{i}\\ \mathbf{q}_{t}\end{bmatrix} \begin{bmatrix}\mathbf{k}_{i}^{T}&\mathbf{k}_{t}^{T}\end{bmatrix} =\begin{bmatrix}\mathbf{q}_{i}\mathbf{k}_{i}^{T}&\mathbf{q}_{i}\mathbf{k}_{t}^{T}\\ \mathbf{q}_{t}\mathbf{k}_{i}^{T}&\mathbf{q}_{t}\mathbf{k}_{t}^{T}\end{bmatrix} \sim \begin{bmatrix}\text{I2I} & \text{T2I} \\ \text{I2T} & \text{T2T}\end{bmatrix}, \label{Eq:attn_map}                                                                                     \\
		 & \mathbf{q}\mathbf{k}^{T}\mathbf{v}= \begin{bmatrix}\mathbf{q}_{i}\mathbf{k}_{i}^{T}&\mathbf{q}_{i}\mathbf{k}_{t}^{T}\\ \mathbf{q}_{t}\mathbf{k}_{i}^{T}&\mathbf{q}_{t}\mathbf{k}_{t}^{T}\end{bmatrix} \begin{bmatrix}\mathbf{v}_{i}\\ \mathbf{v}_{t}\end{bmatrix} = \begin{bmatrix}\mathbf{q}_{i}\mathbf{k}_{i}^{T}\mathbf{v}_{i}+ \mathbf{q}_{i}\mathbf{k}_{t}^{T}\mathbf{v}_{t}\\ \mathbf{q}_{t}\mathbf{k}_{i}^{T}\mathbf{v}_{i}+ \mathbf{q}_{t}\mathbf{k}_{t}^{T}\mathbf{v}_{t}\end{bmatrix}. \label{Eq:attn_result}
	\end{align}
	Since the upper and lower rows in~\cref{Eq:attn_result} correspond to the
	resultant image and text latents, respectively, this notation choice naturally
	reflects the underlying behavior of concatenated attention. Within this generalized
	framework, our empirical findings suggest that I2I and T2I blocks can serve
	roles similar to the conventional self- and cross-attention patterns in U-Net architectures.

	\subsection{Block-wise attention patterns}
	\begin{figure}
		\centering
		\includegraphics[width=\linewidth]{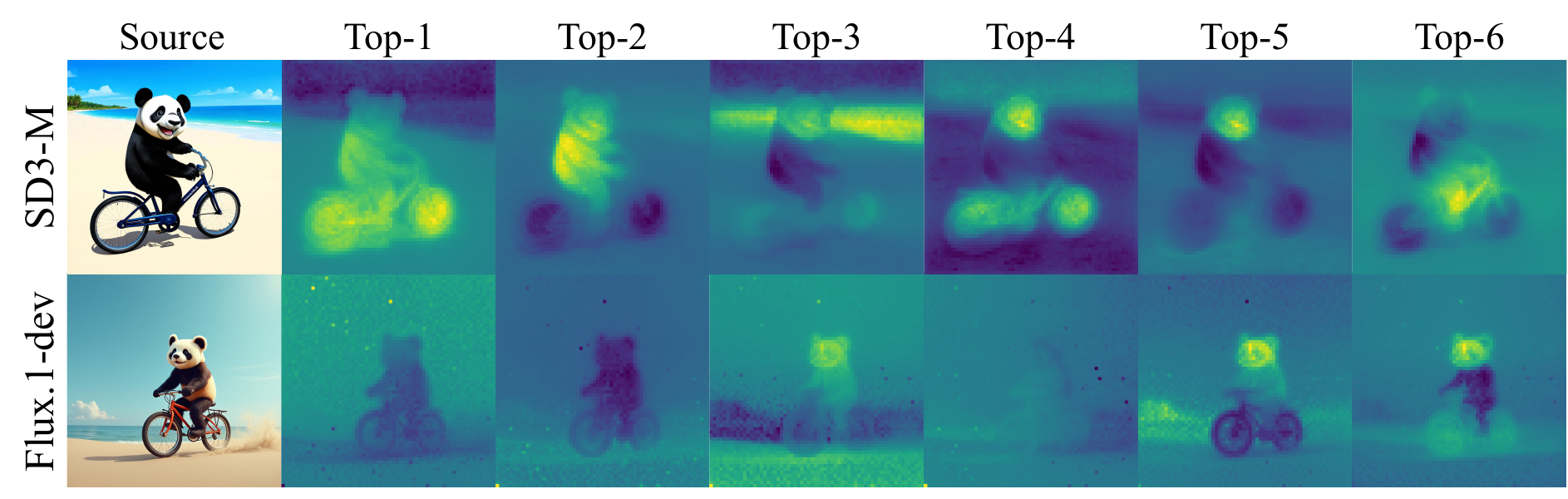}
		\caption{Visualizing the top-6 principal components extracted from I2I
		attention blocks, averaged across all timesteps and transformer blocks. Like
		U-Net's self-attention layers, these components reveal underlying spatial and
		geometric patterns.}
		\label{Fig:pca}
		\vspace{-3mm}
	\end{figure}
	\label{Sec:attention-analysis-block-wise}
	\noindent
	\textbf{I2I \& T2T.} I2I and T2T are unimodal portions of the attention map. As
	demonstrated by PCA analysis in~\cref{Fig:pca}, the I2I block is analogous to
	self-attention in U-Net architectures, capturing spatial layout and geometric information.
	In contrast, T2T blocks predominantly manifest as identity matrices, with
	attention weights notably weakening after the occurrence of meaningful tokens
	(\textit{e.g.,} EOS).

\begin{figure}
		\centering
		\includegraphics[width=\linewidth]{
			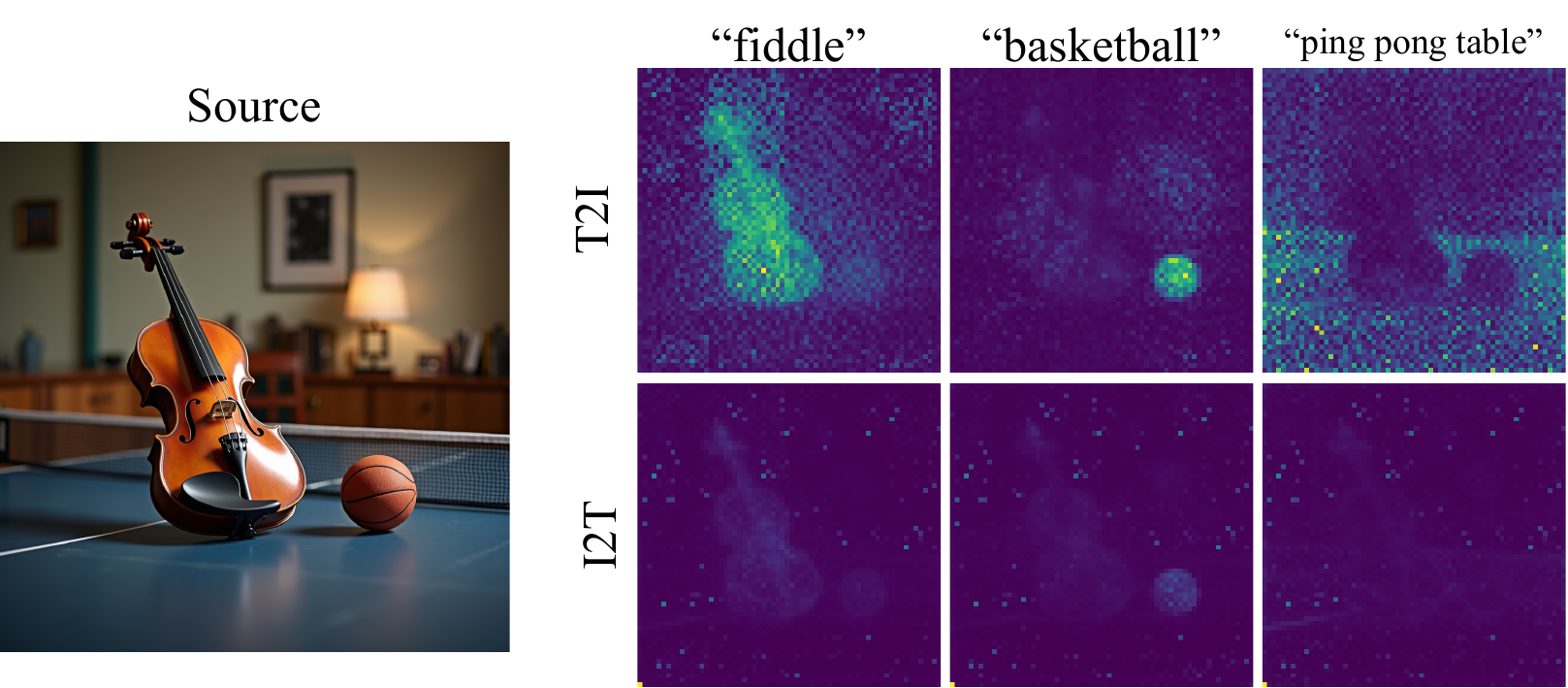
		}
		\caption{Comparison of Flux.1-dev's T2I and I2T attention blocks averaged across
		all timesteps and transformer blocks. T2I blocks show stronger localization
		due to their unconstrained attention distribution compared to I2T's row-wise
		competition, making them more effective source for localized editing.}
		\label{Fig:t2ivsi2t}
		\vspace{-3mm}
	\end{figure}

	\noindent
	\textbf{T2I \& I2T.} T2I and I2T blocks represent cross-modal interactions, where
	T2I influences subsequent image latents, and I2T affects text representations
	- with I2T being a novel addition not present in U-Net architectures. These blocks
	are essential for precise local editing as they encode token-specific image-region
	correspondences, generating binary masks for precise local blending. While both
	blocks can be utilized to obtain these masks, we find T2I more effective (\cref{Fig:t2ivsi2t}).
	Specifically, due to the row-wise softmax operation in I2T blocks, where the rows
	correspond to image tokens, there is inherent competition among these tokens
	since their attention weights must sum to 1. This constraint dilutes the
	attention signals, limiting their ability to represent strong or broad influences.
	In contrast, T2I’s structure allows multiple image regions to retain high attention
	values simultaneously. Additionally, for the SD3 model series using both CLIP and
	T5 encoders, we observe that T5-derived attention maps typically exhibit
	higher precision than CLIP.

	Based on these analyses, we observe that I2I blocks are crucial for preserving
	identity, and T2I blocks are effective in obtaining attention masks for
	precise targeted local editing. We also validate the relative importance of these
	four sub-blocks in~\cref{Fig:sup_bear_to_null_analysis}, which highlights the
	effectiveness of the I2I block in preserving original image attributes. Furthermore,
	visualizations of T2I attention in~\cref{Fig:sup_attention_maps_sd3},~\cref{Fig:sup_attention_maps_sd3.5-M},~\cref{Fig:sup_attention_maps_sd3.5-L},~\cref{Fig:sup_attention_maps_flux}
	suggest that, during full attention's modality mixing, rotary positional embeddings
	(RoPE)~\cite{su2024roformer} and residual connections help each modality retain
	its distinctive characteristics at designated position throughout transformer blocks.
	This supports and justifies our block-based interpretation and editing. Due to
	space constraints, we present extensive additional findings in~\cref{sec:sup_block_wise},
	to which we refer readers for further details.

	\begin{figure}[t]
		\centering
		\includegraphics[width=\linewidth]{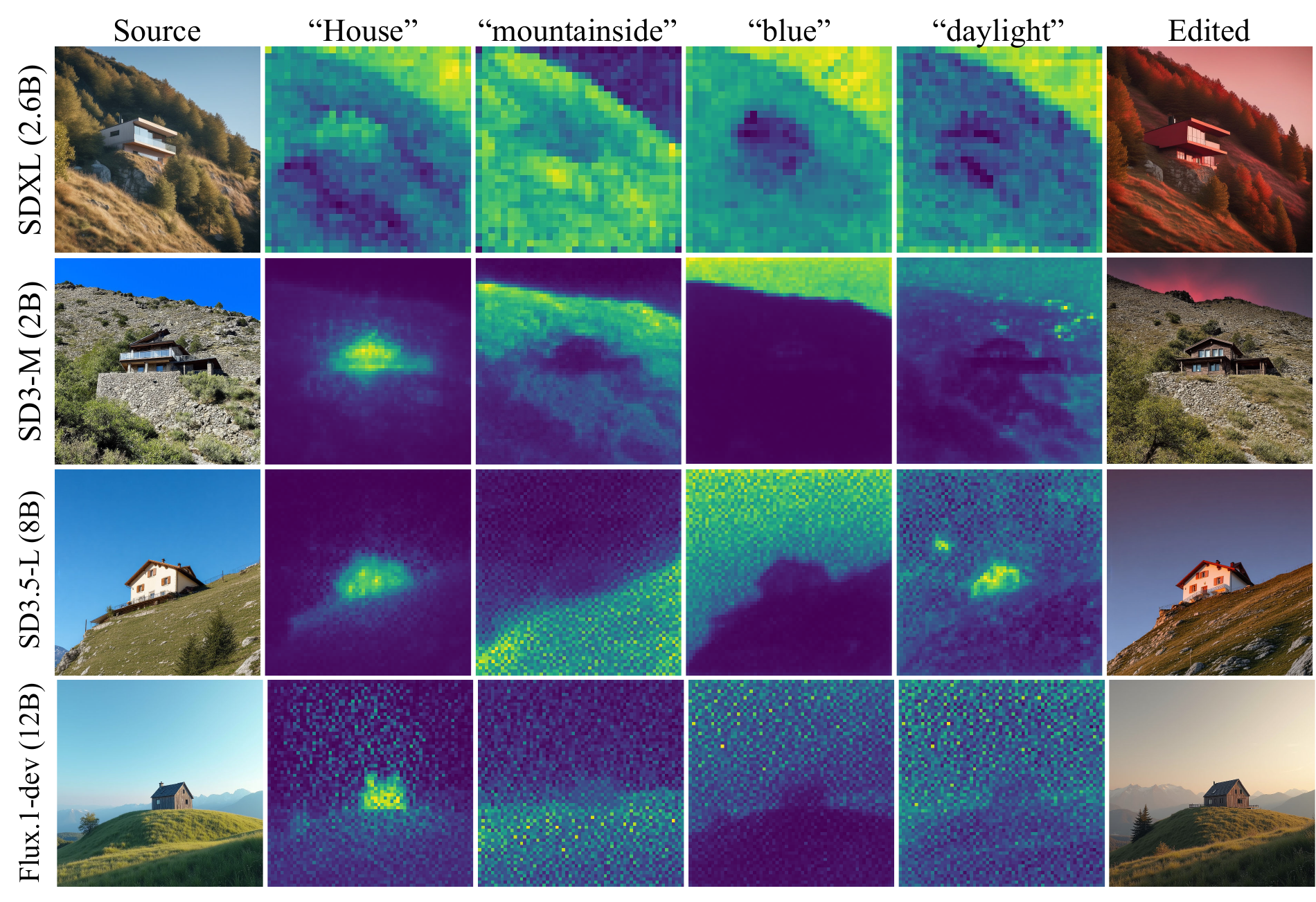}
		\caption{Analysis of T2I attention blocks from SDXL and various MM-DiT
		models using the prompt ``House on a mountainside under a clear blue
		daylight sky'' (averaged across timesteps and transformer blocks). MM-DiT models
		show better localized attention compared to U-Net architectures, though
		attention noise increases with model scale. When editing ``blue daylight" to
		``red dawn" by replacing entire attention blocks, MM-DiT's precise attention
		enables clean edits, while SDXL applies the change globally.}
		\label{Fig:attention_comparison}
		\vspace{-3mm}
	\end{figure}

	\subsection{Mitigating noisy attention maps}
	\label{Sec:attention-analysis-handling}
	\begin{figure}
		\centering
		\includegraphics[width=\linewidth]{
			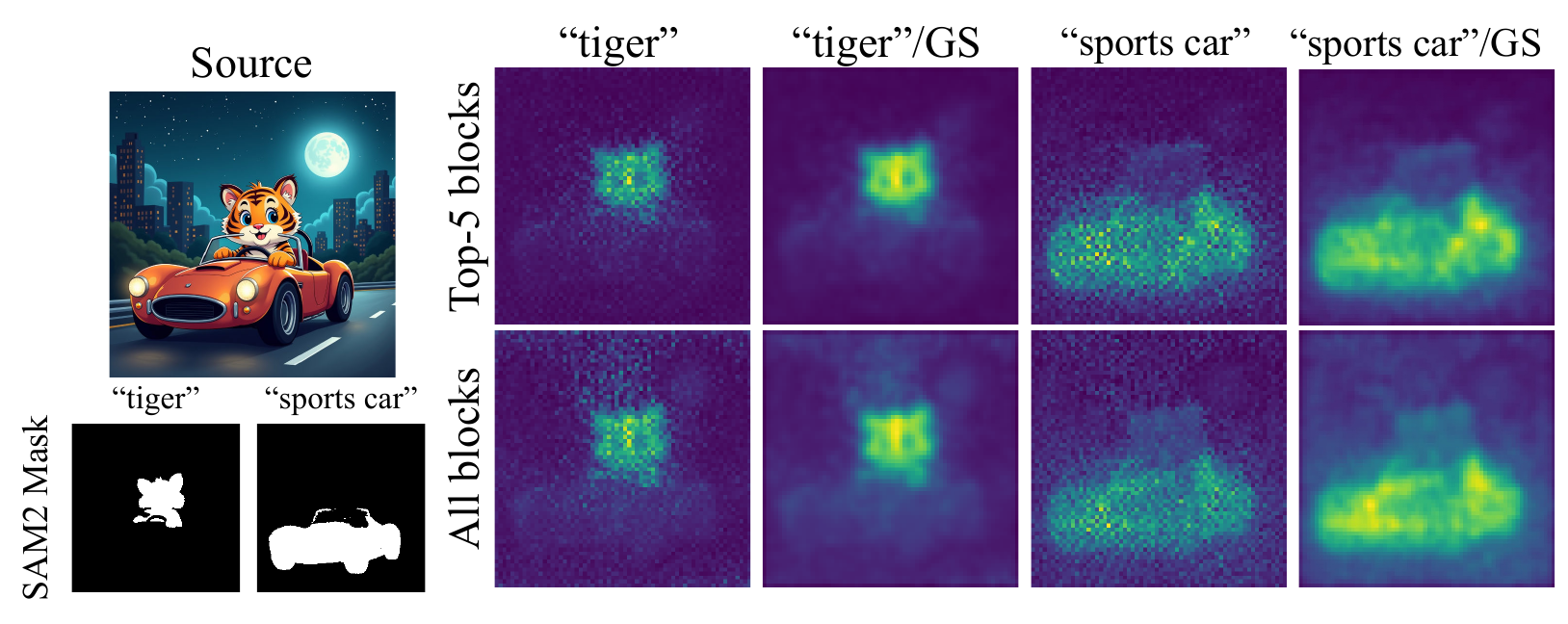
		}
		\caption{Visualization of Flux.1-dev's T2I block, averaged across timesteps,
		comparing masks from top-5 selected blocks (upper) versus all blocks (lower),
		both with and without smoothing. Block selection significantly reduces noise,
		while smoothing refines boundary transitions. The lower left displays Grounded
		SAM2's masks used as ground truth for top-5 block selection.}
		\label{Fig:block_inspection-qual}
		\vspace{-2mm}
	\end{figure}
	We compare cross-attention maps (T2I blocks for MM-DiTs) across various models
	in~\cref{Fig:attention_comparison}. While MM-DiT architectures exhibit significantly
	improved attention localization, we observe an interesting phenomenon: as MM-DiT
	model size grows, the attention maps become increasingly noisy. Directly employing
	these noisy, discontinuous attention maps as local blending masks introduce
	visible artifacts in generated images. This issue aligns with recent findings by~\cite{darcetvision},
	which report similar noise artifacts emerging in ViTs at larger model and
	dataset scales. While they propose to add a register (additional tokens) to remove
	these artifacts, it is impractical here due to the prohibitive cost of
	retraining large MM-DiT models. Thus, we propose two practical solutions: (1) selecting
	optimal transformer blocks that inherently produce more precise and less noisy
	attention maps, and (2) applying Gaussian smoothing to smooth mask boundaries
	and reduce artifacts.

\begin{figure}
		\centering
		\includegraphics[width=\linewidth]{
			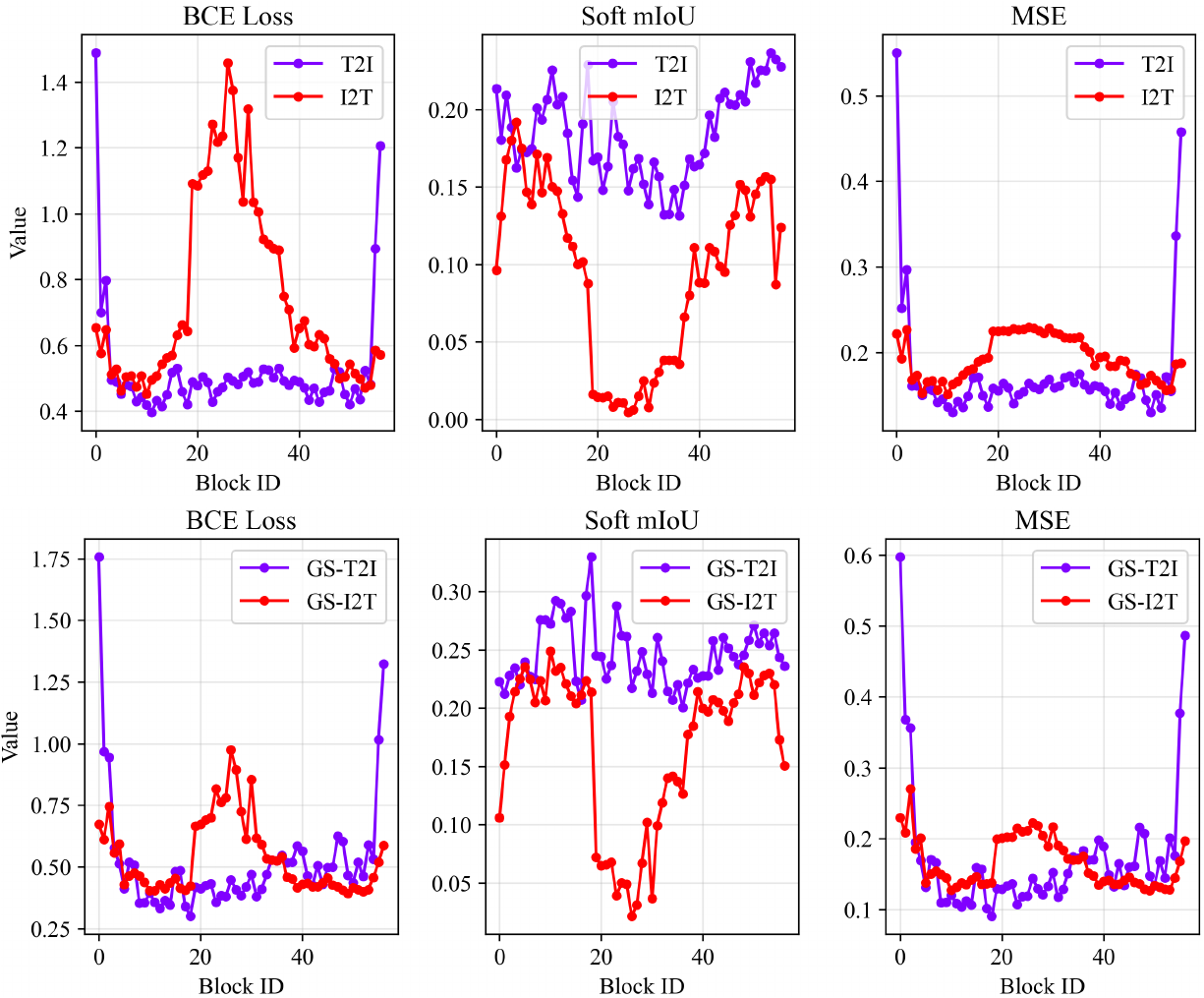
		}
		\caption{Transformer block analysis of Flux.1-dev using Binary Cross Entropy
		Loss, Soft mIoU, and MSE, with Grounded SAM2 predictions as ground truth. Scores
		are shown without (upper) and with (lower) Gaussian smoothing, with all
		evaluation metrics showing notable improvements after smoothing. T2I shows higher
		accuracy than I2T across most blocks. Top-5 ranked T2I attention blocks are
		exclusively used for generating local edit masks. Additional model results are
		provided in~\cref{sec:sup_effective_blocks}.}
		\label{Fig:block_inspection}
		\vspace{-3mm}
	\end{figure}

\begin{figure}[t]
		\centering
		\vspace{1mm}
		\includegraphics[width=\linewidth]{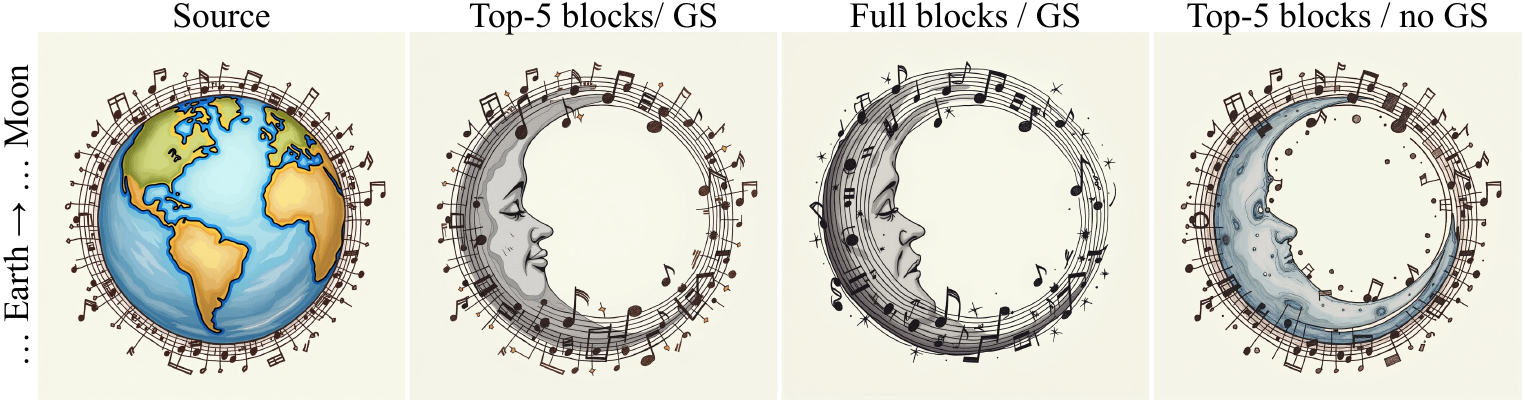}
		\caption{Local image editing results using masks derived from the top-5
		transformer blocks with Gaussian smoothing. Employing only the top-5 blocks reduces
		noise in the attention masks, effectively preserving untouched regions (\textit{e.g.},
		music notes). Applying Gaussian smoothing further reduces boundary artifacts.}
		\label{Fig:block_edit_result}
		\vspace{-4mm}
	\end{figure}

	For systematic evaluation, we sampled 100 prompts from the PARTI prompts~\cite{yuscaling},
	a widely used prompt set for benchmarking, and we created ground truth masks using
	Grounded SAM2~\cite{ravi2024sam2, liu2023grounding} for grounded segmentation tasks,
	as shown in~\cref{Fig:block_inspection-qual}. Intuitively, effective blocks
	should yield attention maps similar to the GT masks. Thus, using standard
	losses and measures (BCE, Soft mIoU, and MSE) against the binary GT masks, we selected
	the top-5 blocks for each model based on the average ranking. Block-wise
	metrics for Flux.1 are presented in~\cref{Fig:block_inspection}, and specific
	rankings are provided in~\cref{Tab:sup_smoothing_comparison}. We emphasize that
	these blocks are \emph{not prompt-specific}, as they were calculated over 100 random
	prompts, and we consistently use the same five blocks throughout the paper. Local
	editing results with and without these selected blocks are presented in~\cref{Fig:block_edit_result}.


	Notably, applying Gaussian smoothing alters the block ranking in larger models,
	suggesting that some noisy blocks still contain meaningful information. In contrast,
	smaller models like SD3-2B, which inherently produce smoother attention maps,
	exhibit stable block rankings regardless of Gaussian smoothing. Unlike previous
	U-Net observations, where down-block features are consistently noisier than up-block
	ones, we observed no strong common trends across MM-DiT variants due to their distinct
	architectural differences. Nonetheless, very early and later blocks consistently
	produced noisier attention maps across all models. Thus, simply excluding these
	noisy blocks and using the identified top-5 blocks consistently led to robust results.
	Another practical advantage of selecting only the top-5 blocks is improved
	computational efficiency. Explicitly computing the full T2I attention map prohibits
	using optimized attention computations such as PyTorch’s dedicated SDPA kernel~\cite{paszke2019pytorch,
	dao2022flashattention,dao2023flashattention2,xFormers2022}. By limiting manual
	full attention computations to just the top-5 blocks, the remaining blocks can
	leverage these optimized kernels, significantly enhancing inference efficiency,
	as further discussed in~\cref{Sec:editing-local-blending}.

	\section{Editing via MM-DiT Attention}
	\subsection{Attention map-based and input projection-based methods}
	\label{Sec:editing-attention-map-based}
	\begin{figure}[t]
		\centering
		\includegraphics[width=0.75\linewidth]{
			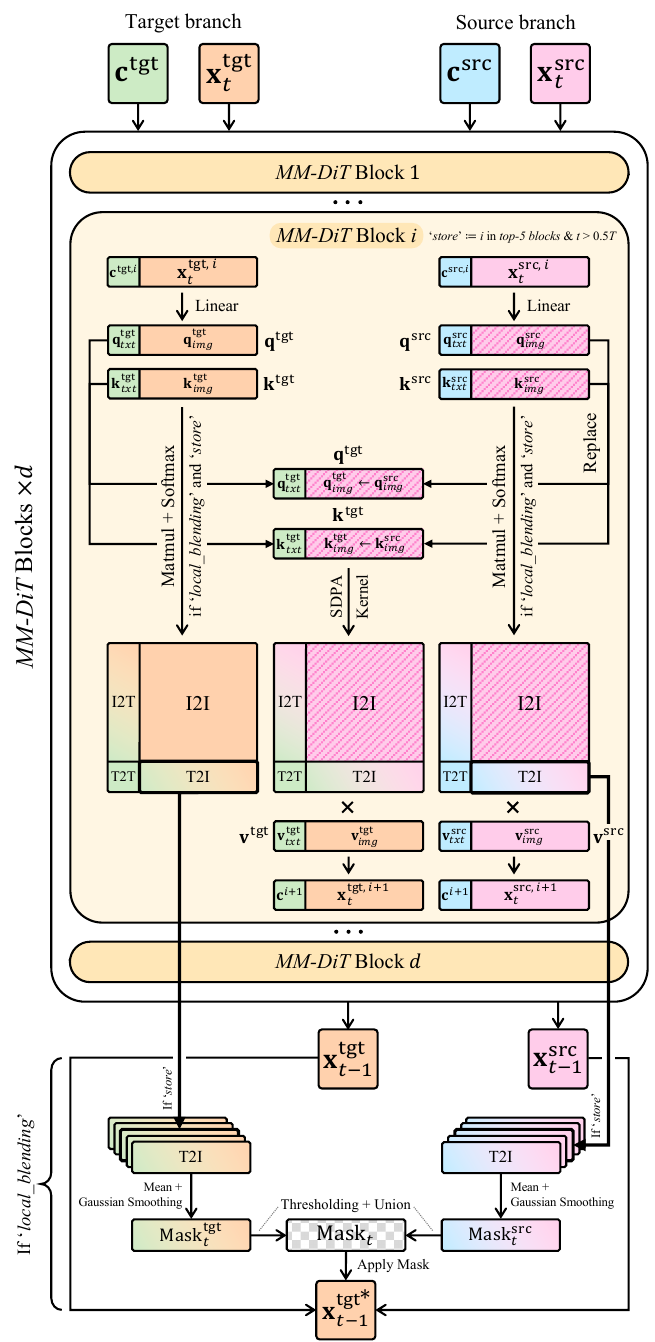
		}
		\caption{Architectural illustration of our method. We replace target (editing)
		branch projections $\mathbf{q}_{i}^{\mathrm{tgt}}$ and $\mathbf{k}_{i}^{\mathrm{tgt}}$
		with their source branch counterparts, $\mathbf{q}_{i}^{\mathrm{src}}$ and
		$\mathbf{k}_{i}^{\mathrm{src}}$. For local blending, we store the T2I
		portion of unmodified attention maps from selected blocks in both branches
		during early timesteps. A binary mask is computed from the union of these attention
		maps after thresholding and used to blend the two latent images. Best viewed
		zoomed in.}
		\label{Fig:main}
		\vspace{-4mm}
	\end{figure}
	\begin{figure}
		\centering
		\includegraphics[width=\linewidth]{
			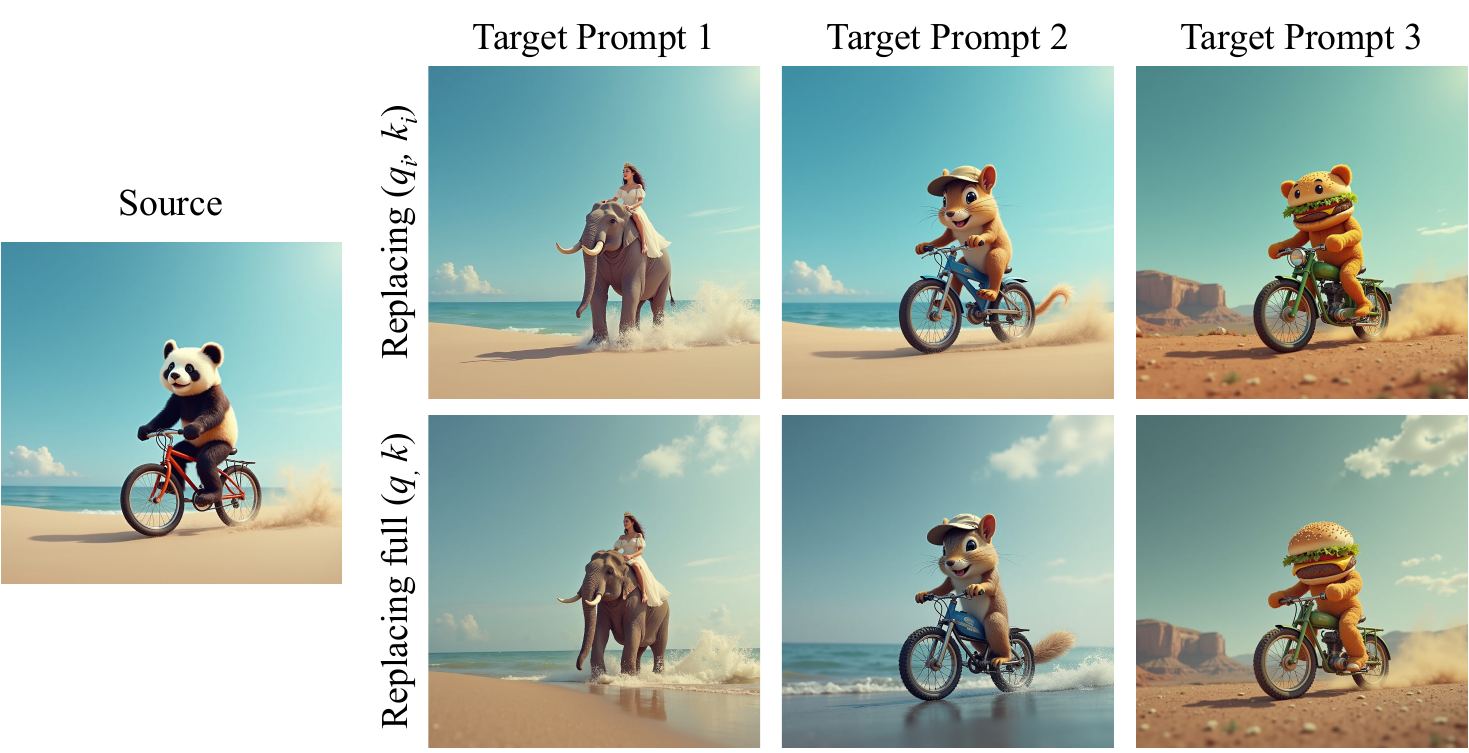
		}
		\caption{Effect of text misalignment on image editing using Flux.1-dev model.
		When editing with highly different prompts (full list in the Appendix), replacing
		only image projections ($\mathbf{q}_{i}$, $\mathbf{k}_{i}$) maintains better
		stability compared to full attention map replacement (all $\mathbf{q}$,
		$\mathbf{k}$).}
		\label{Fig:qkvsqiki}
		\vspace{-3mm}
	\end{figure}

	\begin{figure}
		\centering
		\includegraphics[width=\linewidth]{
			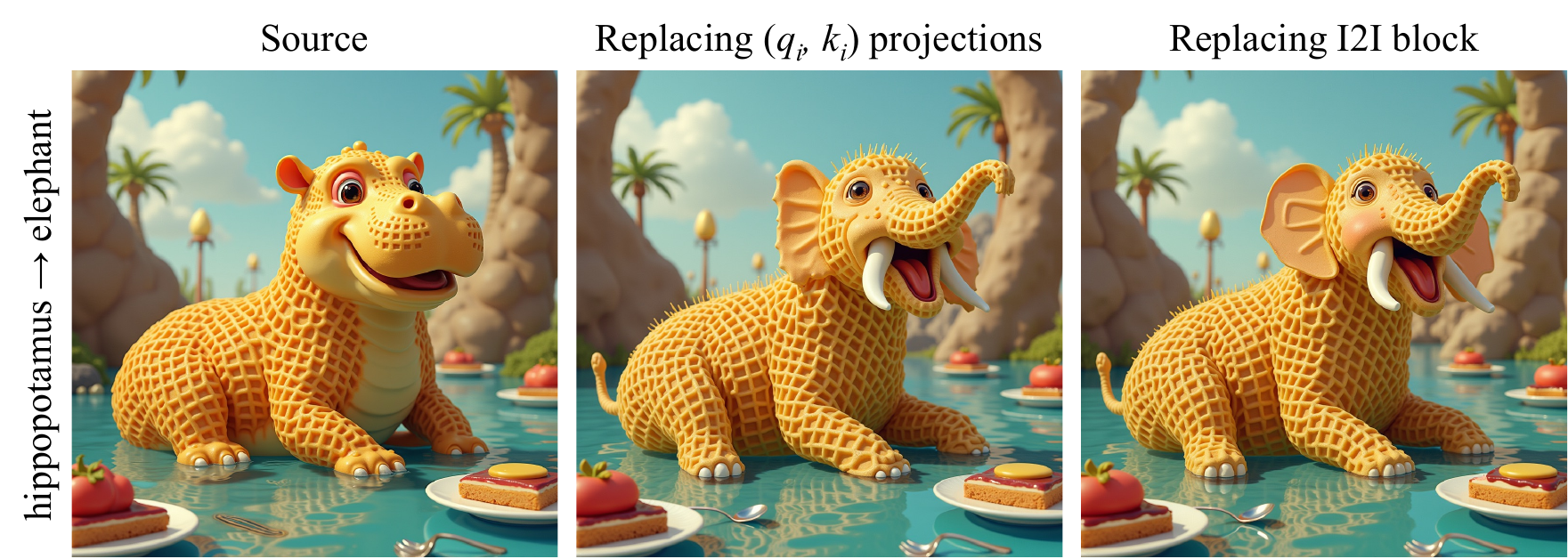
		}
		\caption{Comparison between replacing $\mathbf{q}_{i}$ and $\mathbf{k}_{i}$
		projections and I2I block replacement. Results are highly similar, with
		projection-based methods being more computationally efficient.}
		\label{Fig:qikivsi2i}
		\vspace{-3mm}
	\end{figure}
	Our analysis reveals that MM-DiT attention components serve roles similar to
	their U-Net counterparts: I2I blocks preserve source attributes (self-attention-like),
	while T2I blocks localize text-image interactions (cross-attention-like). Inspired
	by previous studies, we demonstrate image editing by substituting the target
	branch’s attention map with the source branch’s during early denoising
	timesteps. For common rectified flow schedulers (Euler’s method), replacing
	attention maps during the initial 20\% of timesteps yields effective results.

	A key consideration is now determining which attention components to replace.
	While substituting the entire attention map (equivalently, full $\mathbf{q}$, $\mathbf{k}$
	projections) is functional, it introduces text projection misalignment.
	Specifically, replacing the entire target attention map causes the text region
	to shift to the source branch’s contexts ($\mathbf{q}^{\mathrm{src}}_{t}\times
	\mathbf{k}^{\mathrm{src}}_{t}$). When this altered attention is multiplied by
	the target branch’s value matrix text tokens ($\mathbf{v}^{\mathrm{tgt}}_{t}$),
	misalignment arises. With T5 embeddings, subtle prompt differences can significantly
	amplify these misalignments, leading to undesirable image shifts (illustrated in~\cref{Fig:qkvsqiki}).
	This establishes a critical constraint: T2T regions must remain unmodified. In-depth
	explanation of this issue can be found in~\cref{Sec:supediting-text-misalignment}.

	As a solution, we propose two alternative approaches: either selectively replacing
	only the I2I attention block, or specifically modifying just the image input
	projections ($\mathbf{q}_{i}$, $\mathbf{k}_{i}$). Based on~\cref{Eq:attn_map},
	these approaches slightly differ, as modifying $\mathbf{q}_{i}$ and
	$\mathbf{k}_{i}$ also influences T2I and I2T interactions. However, as shown
	in~\cref{Fig:qikivsi2i}, results from both methods are nearly identical.
	Importantly, modifying at the input projection level ($\mathbf{q}_{i}$, $\mathbf{k}
	_{i}$) is computationally more efficient since it enables optimized scaled dot
	product attention (SDPA) kernels. In contrast, replacing the I2I block
	disables SDPA, requiring manual attention calculation and careful switching before
	softmax to maintain proper attention distributions. This absence of SDPA can
	increase the computational time by up to 3$\times$ (\cref{Tab:architecture_details}).

	\begin{algorithm}
		[t] \DontPrintSemicolon \SetInd{0.38em}{0.38em} \scriptsize \SetAlCapFnt{\footnotesize}
		\SetAlCapNameFnt{\footnotesize}
		\caption{\textbf{Prompt-based Image Editing with MM-DiT}}
		\label{alg:synthetic-image-editing} \SetKwInOut{Parameters}{Parameters} \Parameters{Source prompt $P^{\mathrm{src}}$, Target prompt $P^{\mathrm{tgt}}$, Attention replace step $\tau$, Local Blending-\{replace step $\eta$, threshold $\theta$\}}
		$x_{T}^{\mathrm{src}}\!\sim\!\mathcal{N}(0,I)$; \quad $x_{T}^{\mathrm{tgt}}\!
		\gets\!x_{T}^{\mathrm{src}}$\; \For{$t = T, T-1, \ldots, 1$}{ \uIf(\tcp*[f]{Local blend}){$local\_blend$ \textbf{and} $t > \eta$}{ $x_{t-1}^{\mathrm{src}},(\{q,k\}_{i,t}^{\mathrm{src}},AttnMap) \!\gets\!\mathrm{DM}(x_{t}^{\mathrm{src}},P^{\mathrm{src}},t)$\; $x_{t-1}^{\mathrm{tgt}}\!\sgets\!\mathrm{DM}(x_{t}^{\mathrm{tgt}},P^{\mathrm{tgt}},t)\bigl[ \{q,k\}_{i,t}^{\mathrm{tgt}}\!\sgets\!\{q,k\}_{i,t}^{\mathrm{src}}\,\text{if}\,t>\tau \bigr]$\; $\text{Mask}\!\gets\!(\,AttnMap>\theta\,)$\; $x_{t-1}^{\mathrm{tgt}}\!\gets\!x_{t-1}^{\mathrm{tgt}}\!\cdot\!\text{Mask}\;+\;x_{t-1}^{\mathrm{src}}\!\cdot\!(1-\text{Mask})$\; } \Else(\tcp*[f]{No local blend}){ $x_{t-1}^{\mathrm{src}},(\{q,k\}_{i,t}^{\mathrm{src}}) \!\gets\!\mathrm{DM}(x_{t}^{\mathrm{src}},P^{\mathrm{src}},t)$\; $x_{t-1}^{\mathrm{tgt}}\!\sgets\!\mathrm{DM}(x_{t}^{\mathrm{tgt}},P^{\mathrm{tgt}},t)\bigl[ \{q,k\}_{i,t}^{\mathrm{tgt}}\!\sgets\!\{q,k\}_{i,t}^{\mathrm{src}}\,\text{if}\,t>\tau \bigr]$\; } }
		\Return
		$(\,\mathrm{Decode}(x_{0}^{\mathrm{src}}),\,\mathrm{Decode}(x_{0}^{\mathrm{tgt}}
		)\,)$\;
	\end{algorithm}
	To this end, we propose a new approach based on $\mathbf{q}_{i}$, $\mathbf{k}_{i}$
	replacement. Unlike prior U-Net-based attention modification methods that
	require a precise token mapping between source and target prompts, limiting their
	effectiveness when prompts differ significantly, our method can handle
	arbitrarily different prompt pairs. This approach effectively realizes our editing
	objective: maintaining base image similarity while incorporating edited prompt
	elements, a capability uniquely enabled by MM-DiT's full-attention architecture.
	By preserving the text region intact, our method eliminates the need for explicit
	token correspondence or specialized mappers between prompts, allowing robust
	performance across diverse editing scenarios.

	\subsection{Local blending}
	\label{Sec:editing-local-blending}

	Using insights from \cref{Sec:attention-analysis-handling}, we extract clear
	attention maps from the top-5 transformer blocks for precise local blending, ensuring
	only targeted regions change while preserving the base image elsewhere. We
	apply binary blending masks derived from these attention maps once per
	timestep after all blocks are processed, following P2P’s method. This
	selective calculation allows optimized SDPA kernels in remaining blocks, maintaining
	inference speed similar to naive batch size-2 inference (SD3-M: 15.2s vs. 14.9s,
	Flux.1-dev: 55.9s vs. 53.7s, SD3.5-L: 50.1s vs. 47.5s on a single A6000 GPU).
	\begin{figure*}
		\centering
		\includegraphics[width=0.9\linewidth]{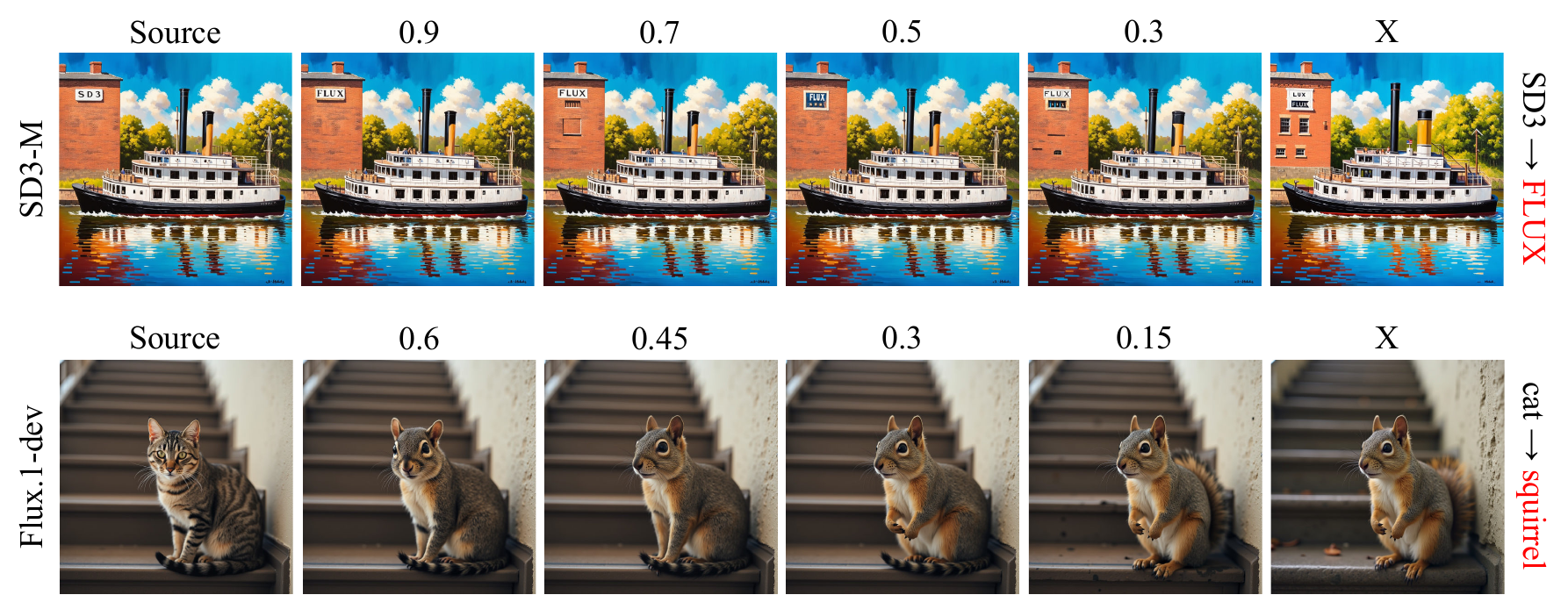}
		\caption{Ablation results on local blending threshold. Fine edits like text modifications
		benefit from higher thresholds, whereas broader transformations like large
		object changes favor lower thresholds. Local blending was applied to the first
		50\% of timestep iterations.}
		\label{Fig:sup_local_blend_abl}
		\vspace{-3mm}
	\end{figure*}

	Our complete pipeline is illustrated in~\cref{Fig:main}, with detailed
	pseudocode provided in~\cref{alg:synthetic-image-editing}. Among the three hyperparameters
	controlling our method, such as masking threshold $\theta$, attention replace timestep
	$\tau$, and local blending stop timestep $\eta$, we find that setting $\tau$ to
	20\% and $\eta$ to 50\% (\textit{i.e.}, $\tau=0.8T, \eta=0.5T$) of total timesteps
	typically achieves robust performance. Thus, the masking threshold $\theta$ remains
	a primary parameter influencing edit granularity. As demonstrated in~\cref{Fig:sup_local_blend_abl},
	higher thresholds effectively enable precise local edits, while lower
	thresholds allow broader modifications. Leveraging MM-DiT’s precise attention localization
	and our attention map curation, the proposed local blending method
	significantly improves edit precision compared to results obtained without
	attention curation (\cref{Fig:sup_local_blend}).

	\subsection{Controlling edit strength via block selection}
	\label{Sec:editing-block-selection} While injecting input projections or
	attention maps into all blocks is straightforward, we observe that later blocks
	have a stronger influence on the final output, producing edited images closely
	resembling the source. While this correlation aligns with our goal, it can
	pose issues for few-step models like Flux.1-schnell, where editing all blocks
	in a single step of a 4-step inference may yield outputs overly similar to the
	base image, regardless of the target prompt. To address this, we replace only the
	first 38 and 30 blocks for Flux.1-schnell and SD3.5-L-Turbo, respectively, while
	replacing every block for non-distilled models. Note that our two block-selection
	strategies (top-5 blocks for obtaining attention maps and blocks for $\mathbf{q}
	_{i}$, $\mathbf{k}_{i}$ projections replacement) are not considered as hyperparameters
	and remain fixed. More details and ablations are available in
	\cref{sec:sup_additional_ablations} and \cref{Fig:sup_block_impact}.

	\begin{figure}
		\centering
		\includegraphics[width=0.97\linewidth]{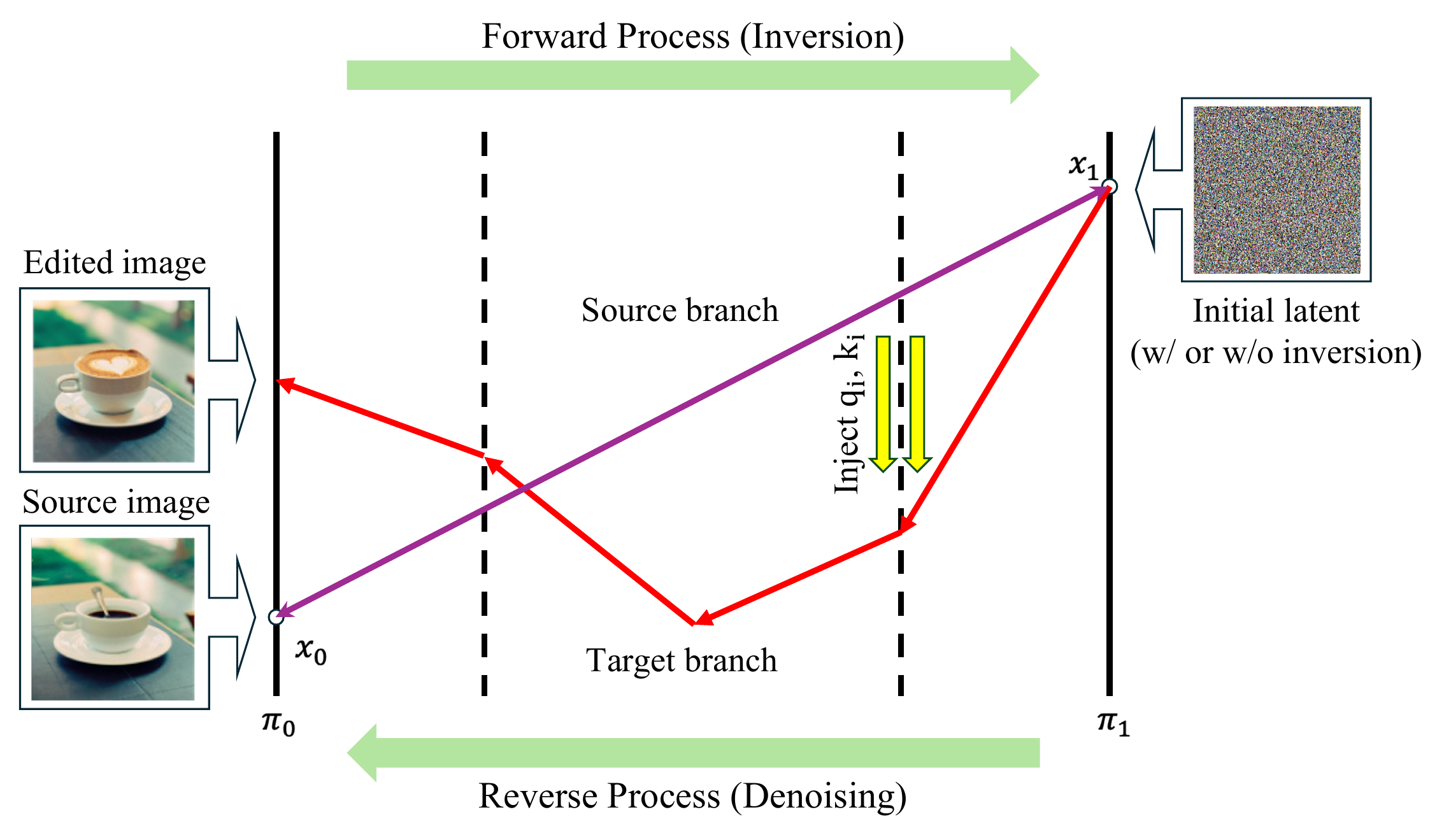}
		\caption{Illustration of our real image editing method given an initial
		latent. Unlike the synthetic image editing case, the source branch is fixed
		as a straight conditional path from the initial noise to the source image, independent
		of model evaluation.}
		\vspace{-4mm}
		\label{Fig:sup_ours_real_image}
	\end{figure}
	\begin{figure}
		\centering
		\includegraphics[width=\linewidth]{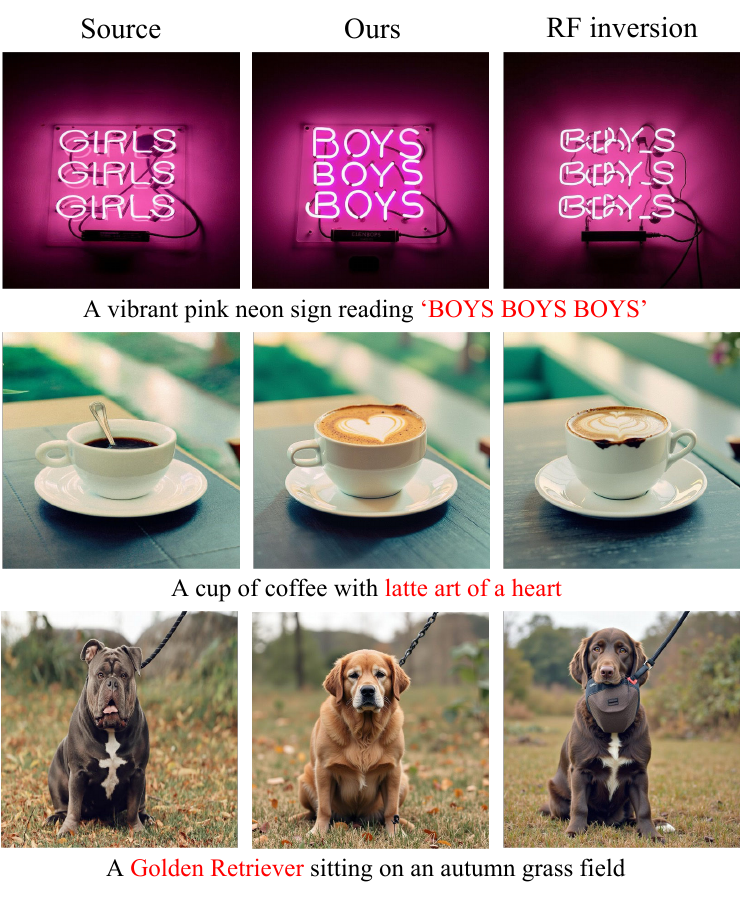}
		\caption{Real image editing comparison between our method and RF inversion. Our
		approach enables more precise local edits while preserving unchanged regions.}
		\label{Fig:realimage}
		\vspace{-0.5mm}
	\end{figure}

	\section{Editing Real Images}
	\label{Sec:real-images} A common approach in existing real image editing methods~\cite{wang2024taming,
	rout2024semantic, avrahami2025stable, dalva2024fluxspace} is to invert the source
	image into an initial noisy latent and then generate the target image from this
	inverted latent using an editing prompt. Our editing technique, which operates
	via $\mathbf{q}_{i}, \mathbf{k}_{i}$ replacement, functions in parallel with
	existing methods and can be readily applied in this scenario. \cref{Fig:sup_ours_real_image}
	illustrates this process, and a preview of our results can be seen in~\cref{Fig:realimage}.
	In~\cref{sec:sup_real_images}, we present further discussions, including quantitative
	and qualitative comparisons, using RF inversion as a baseline. Generally, the
	performance of real image editing methods depends heavily on inversion quality,
	as the initial latent affects background consistency and overall image quality.
	While this dependency also exists for our method, the impact of the initial latent
	is partially mitigated by continuously injecting input projections from the noisy
	source image and performing local blending to preserve unaffected regions.
	Consequently, our method can be used without inversion, starting from random
	noise, offering a trade-off between computational efficiency and performance.

	\section{Experiments}
	\label{Sec:experiments}
	\begin{table}[t]
		\centering
		\caption{Quantitative comparison with different baselines. While higher CLIP
		scores indicate better target prompt alignment, lower LPIPS scores do not
		necessarily reflect edit quality since minimal edits yield the lowest scores.
		Our method well balances source content preservation and target prompt matching,
		whereas fixed seed generation often produces entirely different images and
		prompt-change produces overly conservative modifications.}
		\label{Tab:quantitative_comparison} \footnotesize
		\begin{tabular}{@{}lcccccc@{}}
			\toprule                                                       & \multicolumn{2}{c}{Fixed Seed} & \multicolumn{2}{c}{Prompt Change} & \multicolumn{2}{c}{Ours} \\
			\cmidrule(lr){2-3} \cmidrule(lr){4-5} \cmidrule(lr){6-7} Model & LPIPS                          & CLIP                              & LPIPS                   & CLIP  & LPIPS & CLIP  \\
			\midrule SD3-M                                                 & 0.594                          & 0.377                             & 0.325                   & 0.344 & 0.380 & 0.359 \\
			SD3.5-M                                                        & 0.584                          & 0.380                             & 0.322                   & 0.349 & 0.456 & 0.374 \\
			SD3.5-L                                                        & 0.557                          & 0.388                             & 0.306                   & 0.363 & 0.418 & 0.377 \\
			SD3.5-L-Turbo                                                  & 0.509                          & 0.373                             & 0.225                   & 0.332 & 0.388 & 0.363 \\
			Flux.1-dev                                                     & 0.579                          & 0.359                             & 0.251                   & 0.311 & 0.369 & 0.339 \\
			Flux.1-schnell                                                 & 0.508                          & 0.380                             & 0.116                   & 0.289 & 0.279 & 0.333 \\
			\bottomrule
		\end{tabular}
		\vspace{-2.5mm}
	\end{table}
	\begin{figure}
		\centering
		\includegraphics[width=\linewidth]{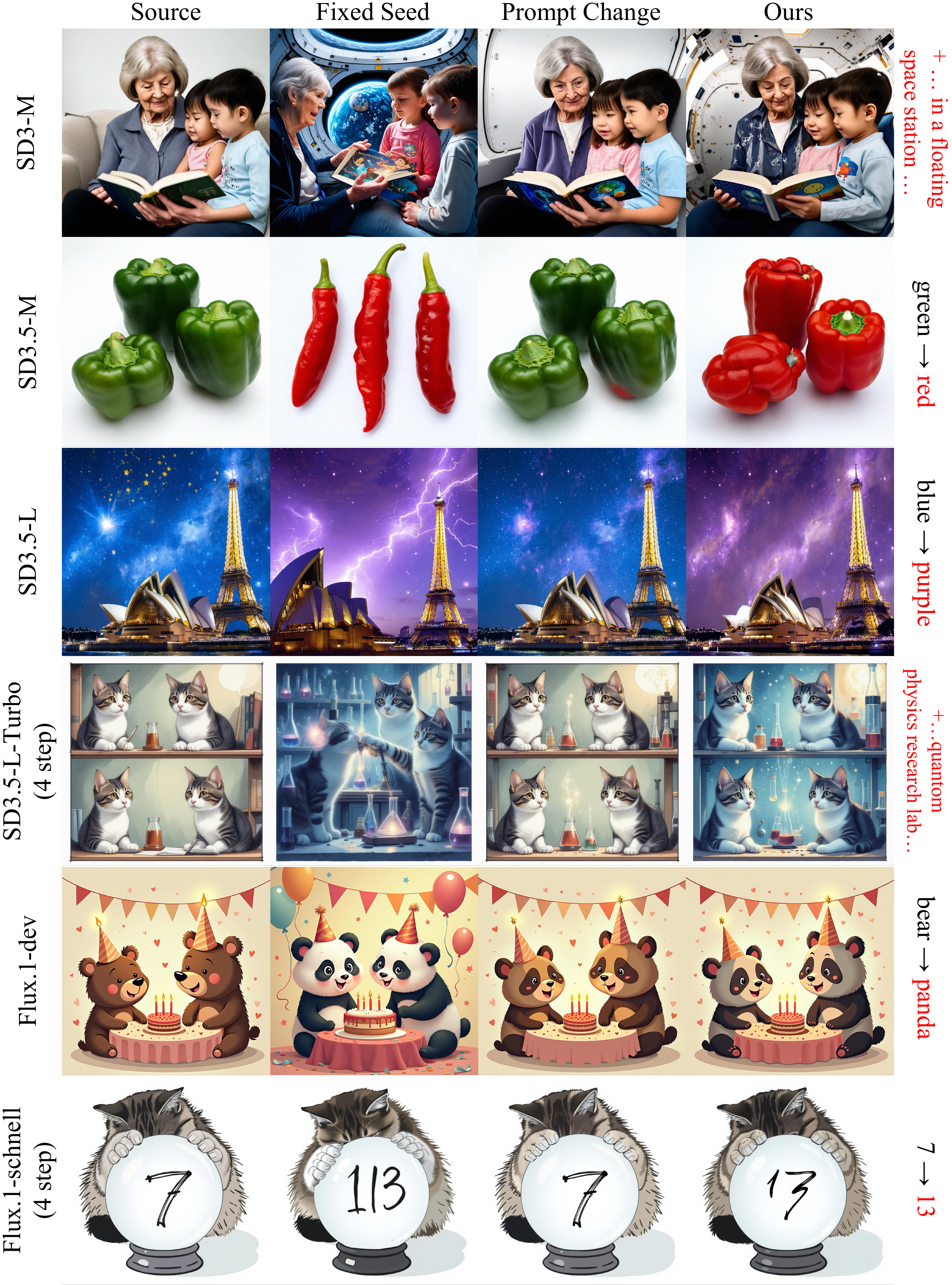}
		\caption{Qualitative comparison of our method with baselines. Our method consistently
		outperforms baseline approaches across diverse editing scenarios.}
		\vspace{-5mm}
		\label{Fig:experiment-comparison}
	\end{figure}
	Due to the lack of dedicated architecture-level editing methods for MM-DiT, we
	validate our approach against two simple baselines: (1) direct generation with
	fixed seeds, and (2) prompt switching after 20\% of timesteps, similar to
	SDEdit~\cite{meng2021sdedit}. We randomly sample 60 diverse prompts from the PARTI
	dataset and use LLM~\cite{Claude2023} to generate edited prompts consisting of
	30 involving simple object or word edits and 30 involving complex changes affecting
	multiple prompt components. We evaluate editing quality using LPIPS~\cite{zhang2018unreasonable}
	and CLIPScore~\cite{hessel2021clipscore}. As these metrics have inherent limitations
	in capturing various types of prompt-based editing quality, we also provide
	user study result in \cref{Fig:user_study_results} and extensive qualitative
	results in \cref{Fig:experiment-comparison}, \cref{Fig:sup_additional_qual1},
	and \cref{Fig:sup_additional_qual2}. All experiments use 28 timesteps, except
	for distilled models (SD3.5-L-Turbo and Flux.1-schnell), which use 4. To ensure
	fair comparisons and avoid hyperparameter tuning, we fix the attention
	replacement step to the first 20\% of iterations and disable local blending for
	the results in \cref{Tab:quantitative_comparison} and \cref{Fig:experiment-comparison},
	though our method can incorporate blending when needed.

	For real-image editing, we observe that editing performance depends more heavily
	on the inversion technique than the architectural-level editing method.
	Nonetheless, we report quantitative results on PIE-Bench~\cite{jupnp} using RF
	inversion as our inversion technique and provide extensive qualitative results.
	Please refer to Appendix~\ref{sec:sup_qualitative_realimage_comparisons} and
	\ref{sec:sup_quantitative_realimage_comparisons} for details. Our method provides
	additional controllability and improves the naive RF inversion baseline.

	\section{Conclusion}
	\label{Sec:conclusion} This work presents a comprehensive analysis of MM-DiT's
	attention mechanism, revealing several underlying insights, including scaling properties
	and text-image interactions. Based on these observations, we develop an efficient
	editing method tailored for MM-DiT architectures that achieves high-quality results
	across various scenarios, including few-step models and real image editing. We
	believe these findings help deepen the understanding of emerging architectures
	and contribute to advancements in controlled image generation and editing.
	Discussions on limitations and future directions can be found in~\cref{sec:sup_limitation}.

	\section{Acknowledgments}
	This work was supported by IITP grant (No. RS-2021-II211343, Artificial Intelligence
	Graduate School Program at Seoul National University) (5\%), IITP grant (No.
	RS-2024-00509257, Global AI Frontier Lab) (65\%), and NRF grant (No. RS-2024-00405857)
	(30\%), funded by the Korea government (MSIT).

	{ \small \bibliographystyle{ieeenat_fullname} \bibliography{main} }
	\clearpage
	\setcounter{page}{1}
	\maketitlesupplementary
	\appendix
	\setcounter{table}{0}
	\setcounter{figure}{0}
	\setcounter{equation}{0}
	\renewcommand{\thefigure}{A\arabic{figure}}
	\renewcommand{\thetable}{A\arabic{table}}
	\renewcommand{\theequation}{A\arabic{equation}}

	\section{Editing Real Images}
	\label{sec:sup_real_images}

	\subsection{Rectified flows}
	\label{sec:sup_rectified_flows} All multimodal diffusion transformers (MM-DiT)
	models discussed in our paper use the setting of Rectified flows~\cite{liuflow}
	for noise scheduling and sampling. Rectified flow presents an approach to
	learn ordinary differential equation (ODE) for transporting between two
	distributions $\pi_{0}$ and $\pi_{1}$(image distribution and standard Gaussian,
	respectively). The key idea is to learn an ODE that follows straight paths connecting
	points drawn from $X_{0}\sim \pi_{0}$ and $X_{1}\sim \pi_{1}$ as closely as possible,
	formulated as follows.
	\begin{equation}
		X_{t}=(1-t)X_{0}+tX_{1}, \quad t \in [0, 1] \label{X_t}
	\end{equation}
	\begin{equation}
		d{Z_t}= v_{t}(Z_{t})dt, \label{rfODE}
	\end{equation}
	\begin{equation}
		v_{t}(x) = \mathbb{E}\left[\dot{X}_{t}\middle| X_{t}= x\right] = \mathbb{E}\left
		[X_{1}-X_{0}\middle| X_{t}= x\right], \label{v_formulation}
	\end{equation}
	$\dot{X}_{t}$ denotes the time differential of ${X}_{t}$. \cref{X_t} defines
	the marginal density at time $t$, corresponding to noise scheduling in diffusion
	models. \cref{rfODE} and \cref{v_formulation} explain the flow connecting each
	sample $Z_{0}\leftarrow \pi_{0}$ and $Z_{1}\leftarrow \pi_{1}$.

	$v_{t}(x;\phi)$ is optimized and evaluated using a neural network through the
	tractable conditional flow matching objective, where $\phi$ represents the trainable
	parameters of the model.
	\begin{align}
		\label{CFM}\mathcal{L}_{\text{CFM}}(\phi) & := \mathbb{E}_{t,X_t,X_1}\left[\|v_{t}(X_{t}|X_{1}) - v_{t}(X_{t}; \phi)\|_{2}^{2}\right], \nonumber \\
		                                          & \text{where}~t \sim \mathcal{U}[0,1], X_{t}\sim p_{t}(\cdot|X_{1}), X_{1}\sim \pi_{1}.
	\end{align}
	\begin{equation}
		v_{t}(x|X_{1}) = \frac{X_{1}- x}{1 - t}, \quad v_{t}(x|X_{0}) = \frac{x-X_{0}}{t}
		, \label{cond_v}
	\end{equation}
	\cref{cond_v}, derived from \cref{X_t} and \cref{v_formulation}, shows that
	the conditional flow follows a straight line to its destination.

	\subsection{RF inversion}
	\label{sec:sup_rf_inversion} Rout et al.~\cite{rout2024semantic} proposed a novel
	inversion framework for RF models that address inversion and editing tasks.
	Inversion is achieved by following a controlled forward ordinary differential
	equation (ODE), establishing a mapping from the real image distribution $\pi_{0}$
	to the standard Gaussian distribution $\pi_{1}$ for the recovery of a noisy latent
	representation from a given real image. Conversely, the controlled reverse ODE
	enables editing by starting with a sample from $\pi_{1}$ and mapping it back to
	$\pi_{0}$, where additional guidance can be applied using target prompts.

	The \textit{controlled forward ODE} maps real image samples from $\pi_{0}$ to standard
	Gaussian $\pi_{1}$ as $t$ progresses from $0$ to $1$. Let
	$\mathbf{x}_{1}\leftarrow \pi_{1}$ denote a sample from the standard Gaussian
	distribution, which serves as a regulation point for inversion. The controlled
	forward vector field $\hat{v}_{t}$ is defined as:
	\begin{equation}
		\label{CF_ODE_forward}\hat{v}_{t}(X_{t}) = v_{t}(X_{t}) + \gamma \big(v_{t}(X
		_{t}\mid \mathbf{x}_{1}) - v_{t}(X_{t})\big), \quad t \in [0, 1].
	\end{equation}
	\begin{equation}
		\label{rfi_inversion_null}v_{t}(X_{t}) = v\big(X_{t}, t, \Phi(``") ; \phi \big
		),
	\end{equation}
	\begin{equation}
		v_{t}(X_{t}\mid \mathbf{x}_{1}) = \frac{\mathbf{x}_{1}- X_{t}}{1 - t},
	\end{equation}

	The controlled vector field $\hat{v}_{t}$ is constructed as a weighted
	interpolation between the unconditional vector field $v_{t}(\cdot)$ with the
	null prompt guidance through text encoder $\Phi$, and the conditional vector field
	$v_{t}(\cdot \mid \mathbf{x}_{1})$, which guides the latent variable toward
	$\mathbf{x}_{1}$ to align better the target distribution $\pi_{1}$. Hyperparameter
	$\gamma$ controls the degree of interpolation between these two fields.

	The \textit{controlled reverse ODE} maps a sample from $\pi_{1}$ back to
	$\pi_{0}$, with $t$ progressing from $1$ to $0$, effectively reversing the
	forward process. Solving this ODE enables reconstruction and editing, with the
	latter guided by a target prompt. The controlled vector field $\hat{v}_{t}$ for
	reverse transformation is defined as:
	\begin{equation}
		\hat{v}_{t}(X_{t}) = v_{t}(X_{t}) + \eta \big(v_{t}(X_{t}\mid \mathbf{x}_{0})
		- v_{t}(X_{t})\big), \quad t \in [0, 1].
	\end{equation}
	\begin{equation}
		\label{rfi_edit_prompt}v_{t}(X_{t}) = v\big(X_{t}, t, \Phi(\text{target
		prompt}) ; \phi \big),
	\end{equation}
	\begin{equation}
		v_{t}(X_{t}\mid \mathbf{x}_{0}) = \frac{X_{t}- \mathbf{x}_{0}}{t},
	\end{equation}

	The formulation is similar to \cref{CF_ODE_forward}, expressed as a weighted
	combination of the unconditional vector field $v_{t}(\cdot)$ with target
	prompt guidance, and the conditional vector field $v_{t}(\cdot \mid \mathbf{x}_{0}
	)$, which incorporates the reference image $\mathbf{x}_{0}$ to align the
	output with the original real image. The interpolation between these fields is
	governed by the hyperparameter $\eta$. Please refer to \cite{rout2024semantic}
	for a detailed theoretical derivation.

	\subsection{Qualitative comparison with other methods}
	\label{sec:sup_qualitative_realimage_comparisons}
	\begin{figure*}
		\centering
		\includegraphics[width=1\linewidth]{
			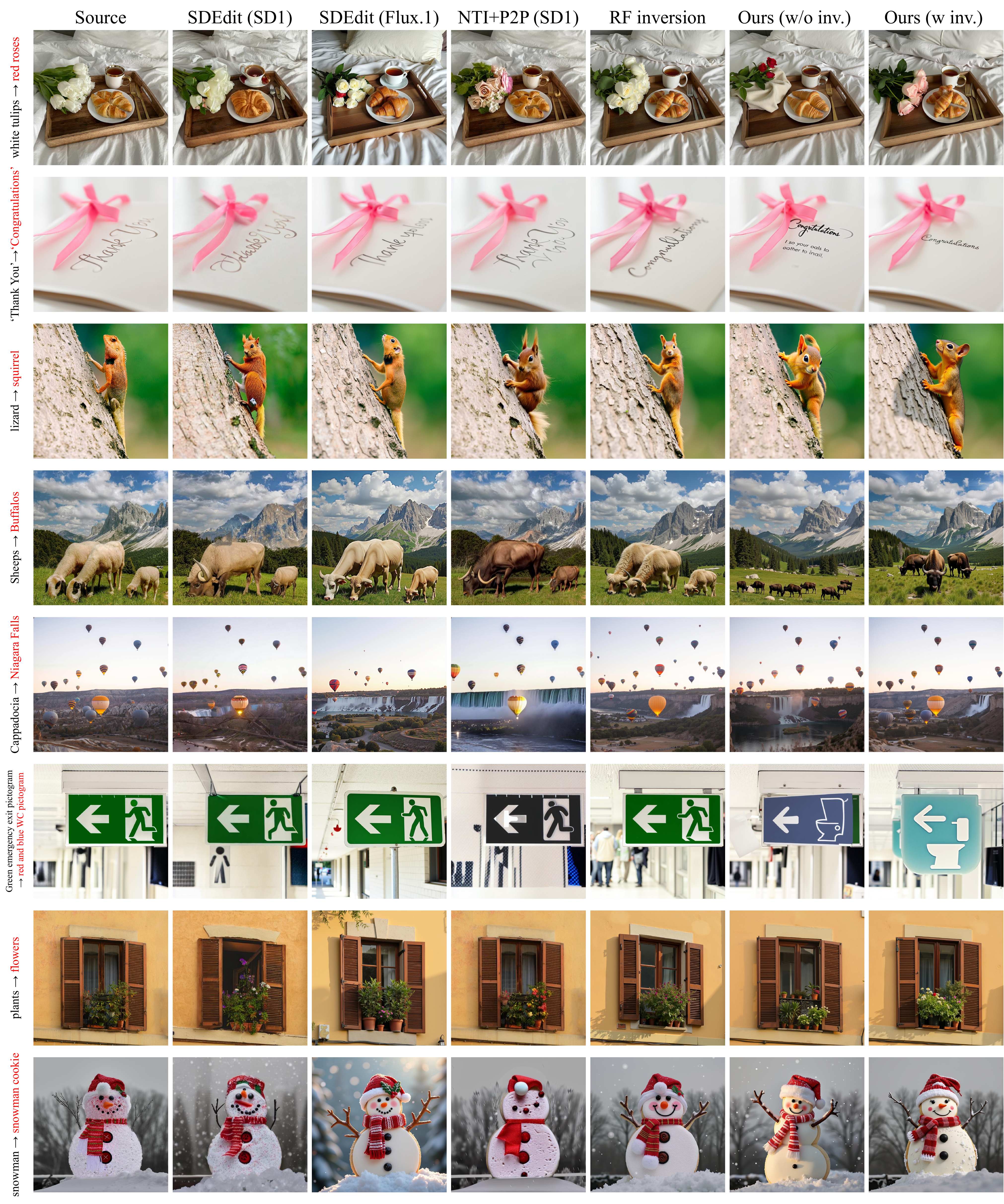
		}
		\caption{Qualitative comparison of real image editing methods. We evaluate using
		diverse real images from Pexels and Pixabay, with their initial captions
		generated by LLM~\cite{GPT4o2024} and subsequently modified for editing
		tasks. Best viewed zoomed in.}
		\label{Fig:sup_real_image_qual}
	\end{figure*}
	\begin{table*}
		[t] \small
		\centering
		\caption{Comparison of diverse image editing methods in PIE-Bench~\cite{jupnp}.
		Best results are in \textbf{bold} and second best are \underline{underlined},
		ranked separately for SD 1.4 and Flux.1-dev methods.}
		\label{Tab:pie_bench_results}
		\begin{tabular}{l|c|c|cccc|cc}
			\toprule \multirow{2}{*}{Method}                          & \multirow{2}{*}{Model / Steps} & Structure             & \multicolumn{4}{c|}{Background Preservation} & \multicolumn{2}{c}{CLIP Similarity} \\
			\cmidrule(lr){3-3} \cmidrule(lr){4-7} \cmidrule(lr){8-9}  &                                & Distance $\downarrow$ & PSNR $\uparrow$                              & LPIPS $\downarrow$                 & MSE $\downarrow$   & SSIM $\uparrow$   & Whole $\uparrow$  & Edited $\uparrow$ \\
			\midrule InstructPix2Pix~\cite{brooks2023instructpix2pix} & SD 1.4 / 50 steps              & 0.057                 & 20.85                                        & 0.158                              & 0.0227             & 0.768             & 23.90             & 21.74             \\
			InstructDiffusion~\cite{geng2024instructdiffusion}        & SD 1.4 / 50 steps              & 0.075                 & 20.31                                        & 0.155                              & 0.0349             & 0.761             & 23.46             & 21.38             \\
			P2P (DDIM-Inv)~\cite{hertz2022prompt}                     & SD 1.4 / 50 steps              & 0.070                 & 17.88                                        & 0.208                              & 0.0219             & 0.717             & \underline{25.31} & \underline{22.57} \\
			Pix2PixZero (DDIM-Inv)~\cite{parmar2023zero}              & SD 1.4 / 50 steps              & 0.062                 & 20.46                                        & 0.172                              & 0.0144             & 0.753             & 23.07             & 20.64             \\
			MasaCtrl (DDIM-Inv)~\cite{cao2023masactrl}                & SD 1.4 / 50 steps              & \underline{0.027}     & 22.19                                        & \underline{0.106}                  & \underline{0.0087} & \underline{0.803} & 24.23             & 21.25             \\
			P2P (PnP-Inv)~\cite{jupnp}                                & SD 1.4 / 50 steps              & \textbf{0.011}        & \textbf{27.28}                               & \textbf{0.054}                     & \textbf{0.0032}    & \textbf{0.853}    & \textbf{25.34}    & \textbf{22.17}    \\
			Pix2PixZero (PnP-Inv)~\cite{jupnp}                        & SD 1.4 / 50 steps              & 0.050                 & 21.56                                        & 0.138                              & 0.0127             & 0.777             & 23.64             & 21.15             \\
			MasaCtrl (PnP-Inv)~\cite{jupnp}                           & SD 1.4 / 50 steps              & 0.024                 & \underline{22.66}                            & 0.087                              & 0.0081             & 0.819             & 24.70             & 21.45             \\
			\midrule RF inversion~\cite{rout2024semantic}             & Flux.1-dev / 28 steps          & \underline{0.026}     & \underline{23.73}                            & \underline{0.144}                  & \underline{0.0065} & \underline{0.769} & \underline{24.56} & \underline{21.59} \\
			Ours ($\theta$=0.2, RF-inv)                               & Flux.1-dev / 28 steps          & 0.054                 & 19.92                                        & 0.204                              & 0.0174             & 0.731             & \textbf{25.43}    & \textbf{22.56}    \\
			Ours ($\theta$=0.5, RF-inv)                               & Flux.1-dev / 28 steps          & \textbf{0.025}        & \textbf{24.79}                               & \textbf{0.126}                     & \textbf{0.0059}    & \textbf{0.804}    & 24.62             & 21.61             \\
			\bottomrule
		\end{tabular}
	\end{table*}
	As discussed in~\cref{Sec:real-images}, our method can be effectively combined
	with inversion techniques. We first obtain the initial latent through
	inversion or sampling from a Gaussian distribution. We then define a conditional
	interpolation path between this initial latent and the image latent using \cref{X_t},
	which we treat as the source branch. During denoising, we simultaneously
	evaluate the model using both the source and target branches, replacing the target
	branch’s input projections with those derived from the source branch. Note
	that the source branch outputs are solely employed to obtain the
	$\mathbf{q}_{i}$ and $\mathbf{k}_{i}$ projections required to update the
	target branch. The remaining outputs from the source branch are disregarded,
	as the interpolation path between the initial latent and source image latent
	is already theoretically defined.

	We provide qualitative comparisons in~\cref{Fig:sup_real_image_qual}, against several
	baseline methods: (1) SDEdit~\cite{meng2021sdedit} based on SD1, (2) SDEdit based
	on Flux.1-dev, (3) Null-text inversion (NTI)~\cite{mokady2023null} with Prompt-to-Prompt~\cite{hertz2022prompt}
	based on SD1, and (4) RF inversion~\cite{rout2024semantic}, along with our results
	both with and without RF inversion. For implementation, we used community implementations
	for SDEdit variants and RF inversion and the official implementation for NTI+P2P.
	SD1-based methods were experimented with default settings (50 timesteps at $512
	\times 512$ resolution), while Flux.1-dev-based methods used 28 timesteps and
	$1024 \times 1024$ resolution. As mentioned in~\cref{Sec:editing-attention-map-based},
	original P2P only allows changes with the same word counts, and it does not support
	changing words like `Cappadocia' to `Niagara Falls' due to different word counts.
	To address this limitation, we manually modified some prompts for NTI+P2P by removing
	spaces between words (\textit{i.e.}, `NiagaraFalls'). All baseline
	hyperparameters were empirically optimized.

	In general, SD1-based methods occasionally show limitations in output quality due
	to the base model's capacity. Compared to SDEdit (Flux.1) and RF inversion,
	our method enables larger changes while naturally preserving unmodified
	regions. When our approach is applied without inversion, results become more
	sensitive to the local blending threshold ($\theta$), requiring higher thresholds
	to effectively maintain targeted regions due to divergence in the denoising sequence.
	In contrast, starting from optimized inverted latents inherently preserves source
	image characteristics, making results less sensitive to threshold values, as lower
	thresholds are already sufficient.

	\begin{figure*}
		\centering
		\includegraphics[width=\linewidth]{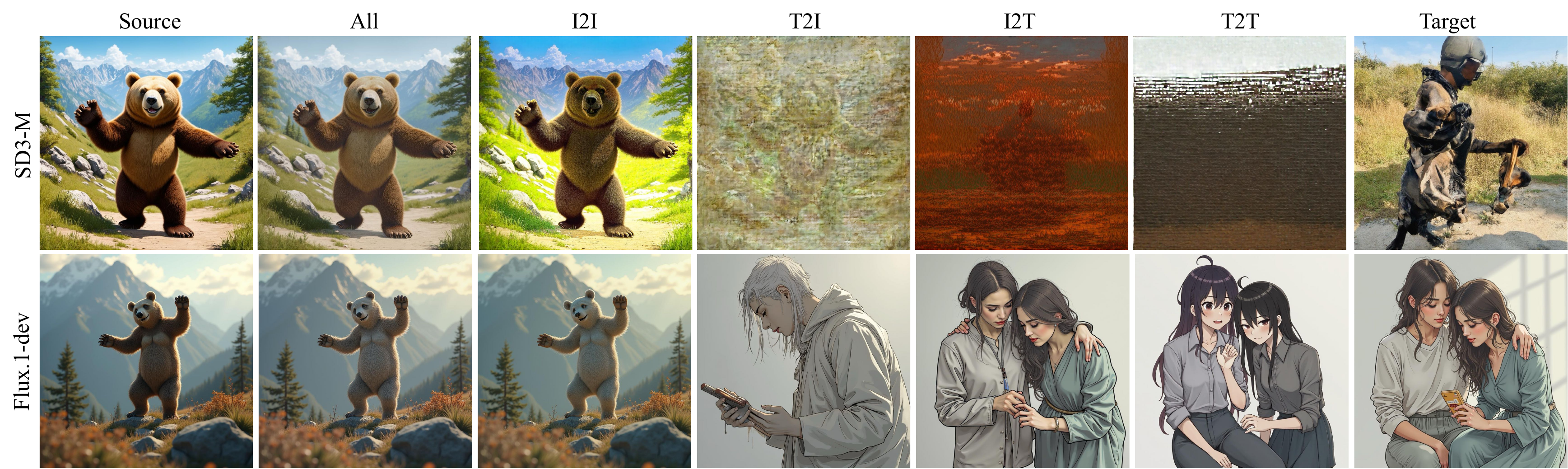}
		\caption{Analyzing attention map components by replacing different portions
		from a source prompt (``a photo-realistic bear dancing in the mountain") to an
		empty target prompt (``''). Results show I2I portions primarily preserve spatial
		layout and geometry, with T2T adding the most negligible impact. Full attention
		map replacement produces the closest match to source image.}
		\label{Fig:sup_bear_to_null_analysis}
	\end{figure*}
	\subsection{Quantitative comparison with other methods}
	\label{sec:sup_quantitative_realimage_comparisons} We evaluate our method on PIE-Bench~\cite{jupnp},
	a benchmark for prompt-based image editing consisting of 700 samples. To maximize
	the model's performance, we compare Flux.1 and SD 1.4-based methods at their native
	resolutions of 1024 and 512, respectively. Our inversion-free method requires careful
	local blending threshold $\theta$ control, so we only perform experiments with
	inverted latents using RF inversion and fixed blending threshold. As seen in \cref{Tab:pie_bench_results},
	our approach improves upon RF inversion with increased controllability via
	$\theta$. While lowering $\theta$ enables broader edits, selecting an
	appropriate $\theta$ allows our method to improve RF inversion in both edit
	quality and image preservation.

	Overall, we observed that Flux-based methods effectively reflect the desired edit
	prompts but demonstrate relatively weaker identity preservation than SD 1-based
	methods. We identify two main reasons behind this:

	\textbf{1) Inversion method}: RF inversion employs first-order Euler methods
	using a controlled vector field derived through dynamic optimal control, interpolating
	between two vector fields: an unconditional vector field guiding images to
	noise, and a controlled vector field that ensures the inverted latent to be closer
	to a ``typical'' latent of Flux's latent distribution. While RF inversion
	performs reasonably well, starting from better inverted latents could potentially
	yield higher scores. As can be seen with the SD1 case, changing the inversion method
	from first-order DDIM to more advanced PnP inversion improves scores throughout.
	Additionally, RF inversion utilizes a controlled interpolation mechanism explicitly
	designed to guide inverted latents toward Flux’s distribution of clean images,
	sometimes producing reconstruction that appear perceptually sharper or cleaner
	than the source images, paradoxically leading to lower reconstruction metrics such
	as PSNR and LPIPS.

	\textbf{2) Dataset characteristics}: PIE-Bench, proposed by~\cite{jupnp}, was mostly
	developed within SD1's capabilities, involving relatively simpler and slightly
	noisy images at SD1's favorable $512\times512$ resolution. When running at Flux.1's
	native $1024\times1024$ resolution, the refinement effect often adversely affects
	identity preservation metrics. Additionally, the benchmark contains relatively
	fewer examples that can properly evaluate MM-DiT's complex and precise control
	capabilities, such as editing text.

	% Considering that PnP~\cite{jupnp} found optimal hyperparameters through
	% extensive grid searches, we found RF inversion’s performance - achieved primarily
	% via first-order Euler steps and controlled interpolation without comprehensive
	% hyperparameter sweeps - to be quite reasonable.
	PnP~\cite{jupnp} achieved strong performance through extensive grid searches
	for optimal hyperparameters. In contrast, RF inversion performs reasonably
	well using only first-order Euler steps and controlled interpolation, without thorough
	hyperparameter tuning. Our visual inspection on edited images revealed many
	high-quality results with RF inversion, and our method enhances them with additional
	controllability. As existing benchmarks tend to use simpler scenes (mostly fitted
	to SD1), we believe evaluating larger models on more complex scenes and tasks (\textit{e.g.},
	text rendering, or high-resolution broader edits) remains an important
	direction.

	\section{Block-wise Attention Patterns}
	\label{sec:sup_block_wise}
	\begin{figure*}
		\centering
		\includegraphics[width=1.0\linewidth]{
			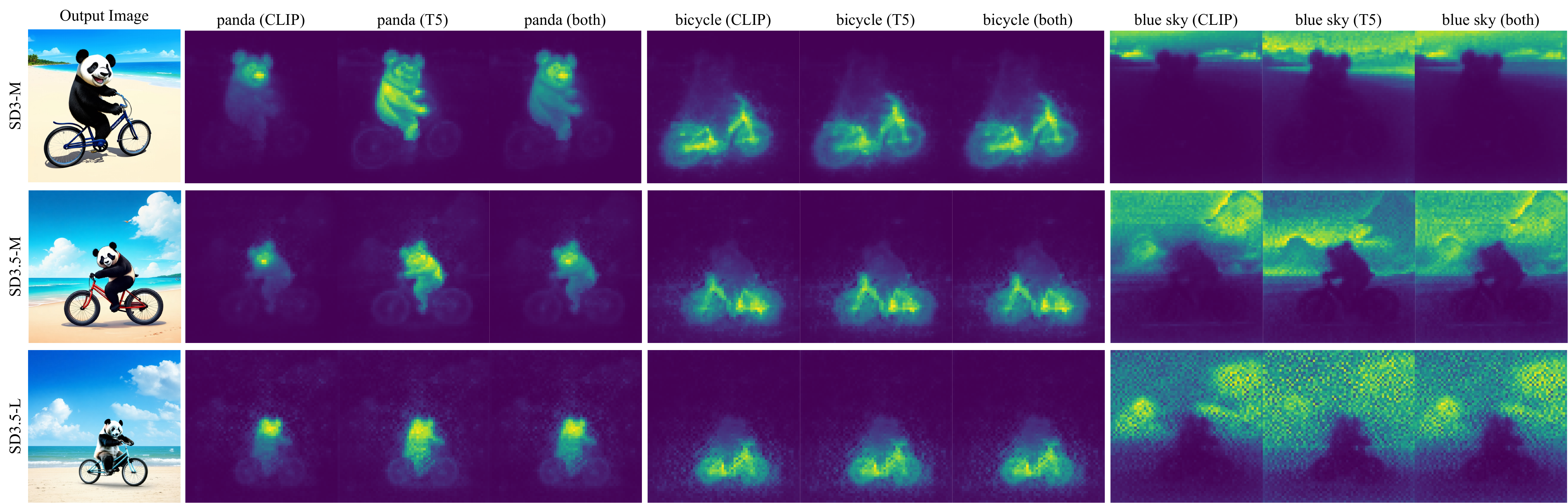
		}
		\caption{Visualization of T2I attention maps separated by text encoder type.
		For SD3 variants using both CLIP and T5 embeddings, we visualize attention
		patterns from CLIP tokens, T5 tokens, and their combination by tracking
		respective token positions.}
		\label{Fig:sup_clipvt5}
	\end{figure*}

	\subsection{Additional discussions on I2I \& T2T blocks}

	As mentioned in the main paper, the I2I block is analogous to self-attention in
	U-Net architectures, effectively capturing spatial layout and geometric
	information. In contrast, T2T blocks primarily manifest as identity matrices, indicating
	strong self-correlation among tokens. To validate the relative importance of
	these sub-blocks, we conducted experiments injecting attention maps from meaningful
	prompts into the empty prompt (``'') branches (\cref{Fig:sup_bear_to_null_analysis}).
	While full attention map transfer produced the closest replication of source images,
	we found that the I2I block alone sufficiently preserves geometric structure, whereas
	T2T has minimal impact.

	To further investigate T2T blocks, we visualize their attention patterns using
	the prompt ``a panda riding a bicycle on the beach under blue sky'' in~\cref{Fig:sup_t2t}.
	For the SD3-M variant, which utilizes 333 tokens (77 CLIP + 256 T5), we observe
	pronounced attention signals around special tokens, particularly at sequence
	boundaries such as start/end tokens and transitions from CLIP to T5 embeddings.
	Similarly, Flux.1-dev, which employs 512 T5 tokens exclusively, also exhibits notable
	attention at prompt endings, with attention weights substantially decreasing
	after meaningful tokens (\textit{e.g.}, EOS). These patterns suggest a focused
	allocation of attention toward semantically relevant token boundaries.
	Additional subtle and noisy patterns within T2T blocks require further
	exploration, which we defer to future work due to T2T blocks' minimal impact
	on current editing scenarios.

	\subsection{Additional discussions on T2I \& I2T blocks}
	We begin by visualizing the T2I and I2T portions of attention maps across
	several model variants (SD3-M:~\cref{Fig:sup_attention_maps_sd3}, SD3.5-M:~\cref{Fig:sup_attention_maps_sd3.5-M},
	SD3.5-L:~\cref{Fig:sup_attention_maps_sd3.5-L}, Flux.1:~\cref{Fig:sup_attention_maps_flux}).
	As discussed in the main paper, we observe spatially and geometrically aligned
	visual patterns when we visualize attention patterns between image tokens and
	specific text tokens. This alignment indicates that each domain preserves its distinct
	characteristics even within the multimodal full attention mechanism. Notably, this
	phenomenon persists in Flux.1's single-branch blocks, where a unified set of weights
	processes concatenated tokens. This observation suggests that the
	architectural choice between dual and single branches does not largely
	compromise the model's ability to maintain domain-specific features. It is also
	worth noting that certain blocks produce extremely noisy attention maps, which
	validates our strategy of utilizing only selected well-defined blocks for
	local blending to achieve more precise local edits.

	\subsection{Comparing CLIP and T5 text encoders}
	Another notable aspect is the use of T5 text encoders alongside CLIP text
	encoders in SD3 series. As shown in~\cref{Tab:architecture_details}, all
	Stable Diffusion 3 series models we tested (SD3-M, SD3.5-M, SD3.5-L) utilize
	three text encoders for the text branch in MM-DiT, concatenating two CLIP text
	embeddings with T5 text embeddings along the sequence dimension, whereas Flux.1
	exclusively uses T5 for the text branch and utilizes CLIP features only as pooled
	embeddings for scale and shift operations. In~\cref{Fig:sup_clipvt5}, we visualize
	CLIP and T5 attention patterns separately. CLIP text encoders generally
	produce denser, more localized attention patterns focused on specific regions.
	In contrast, T5 encoder generates more spread-out attention patterns that appear
	more contextual, sometimes extending to related concepts (\textit{e.g.}, ``blue
	sky" attention spreading to ocean regions due to shared blue attributes). In
	SD3 / 3.5 architectures, we naturally utilize attention maps from both CLIP and
	T5 text token positions when aggregating T2I blocks to generate local blending
	masks.

	\subsection{Detailed explanation of token misalignment}
	\label{Sec:supediting-text-misalignment} In~\cref{Sec:editing-attention-map-based},
	we discussed how changing the entire attention map can lead to misalignment
	with the value matrix. Here, we explain this using the example prompt ``a
	panda riding a bicycle on the beach under blue sky''. The CLIP tokenizer produces
	[`a', `panda', `riding', `a', `bicycle', `on', `the', `beach', `under', `blue',
	`sky'], while the T5 tokenizer yields [`', `a', `pan', `d', `a', `riding', `',
	`a', `bicycle', `on', `the', `beach', `under', `blue', `sky']. When editing with
	a similar prompt ``a dragon riding a bicycle on the beach under blue sky", CLIP
	tokenization simply requires mapping `panda' to `dragon' as they are both single
	tokens. However, in T5, we need to create a mapper that maps all three tokens
	(`pan', `d', `a') to a single `dragon' token. While P2P handles such cases by defining
	explicit token mappings, this approach becomes challenging with drastically
	different prompts like ``a princess with a crown riding an elephant on the
	beach under blue sky", where determining appropriate token correspondences is
	non-trivial. This limitation becomes more pronounced in larger models with longer,
	more descriptive prompts and T5 tokenization. As shown in~\cref{Fig:qkvsqiki},
	while naive attention map replacement leads to undesired changes due to these
	token misalignments, our approach of modifying only image tokens naturally circumvents
	this limitation by keeping text token projections intact.

	\section{In-depth Analysis of Transformer Blocks}
	\begin{table}
		\centering
		\caption{Top-5 block indices calculated using rankings of BCE loss, Soft mIoU,
		and MSE. Results shown with and without Gaussian smoothing across different model
		variants.}
		\label{Tab:sup_smoothing_comparison} \small
		\begin{tabular}{@{}lcc@{}}
			\toprule Model & w/o Gaussian Smoothing & w. Gaussian Smoothing \\
			\midrule SD3-M & [7, 8, 5, 4, 9]        & [7, 8, 5, 4, 9]       \\
			SD3.5-M        & [7, 8, 5, 9, 6]        & [7, 9, 8, 5, 10]      \\
			SD3.5-L        & [18, 16, 29, 21, 14]   & [18, 21, 20, 24, 16]  \\
			Flux.1-dev     & [11, 50, 18, 13, 10]   & [18, 17, 12, 14, 11]  \\
			\bottomrule
		\end{tabular}
	\end{table}
	\begin{figure}
		\centering
		\includegraphics[width=\linewidth]{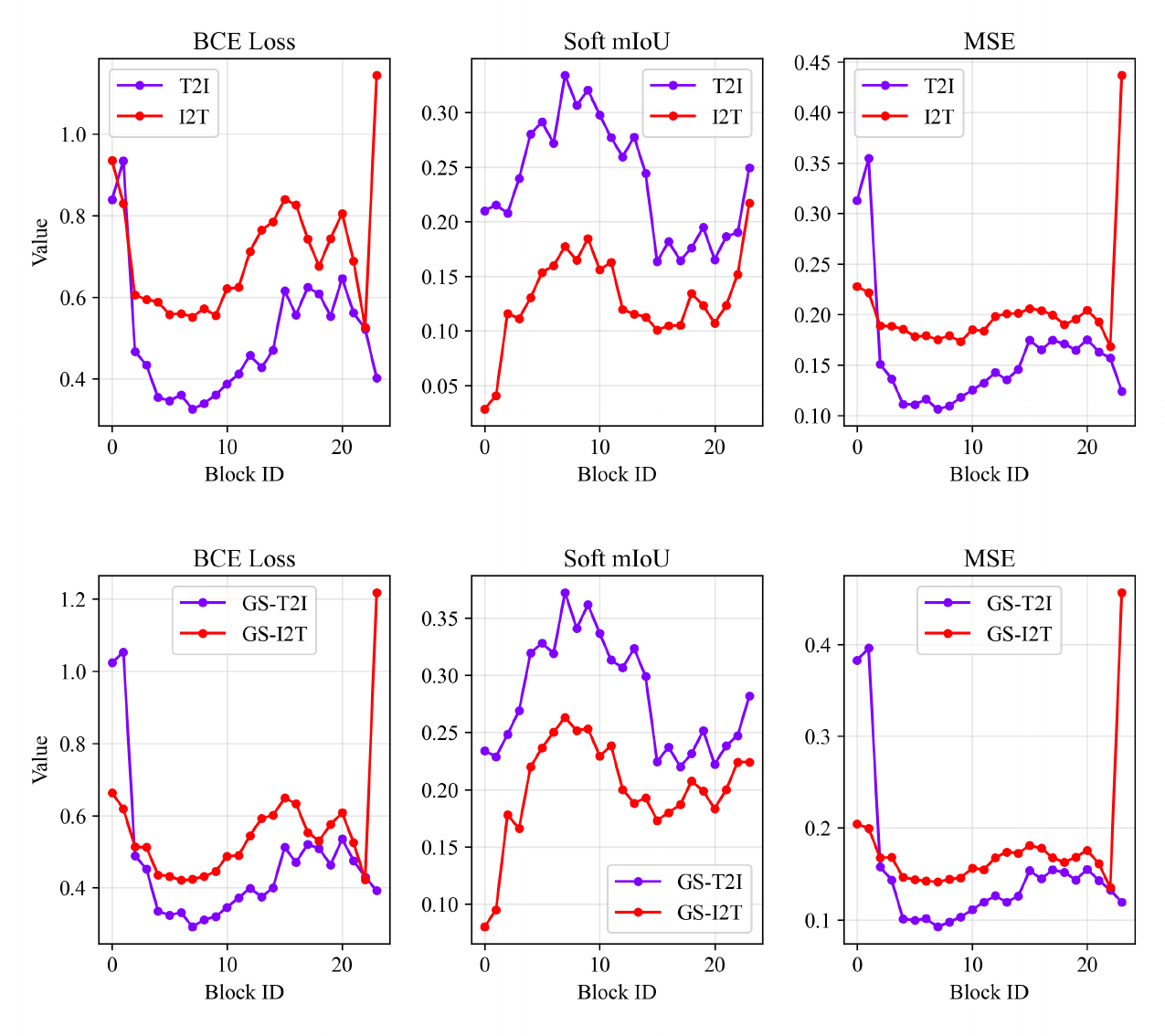}
		\caption{Transformer block analysis of SD3-M using Binary Cross Entropy Loss,
		Soft mIoU, and MSE, with Grounded SAM2 predictions as ground truth. Scores
		are shown without (upper) and with (lower) Gaussian smoothing.}
		\label{Fig:sd3-M-block-inspection}
	\end{figure}
	\begin{figure}
		\centering
		\includegraphics[width=\linewidth]{
			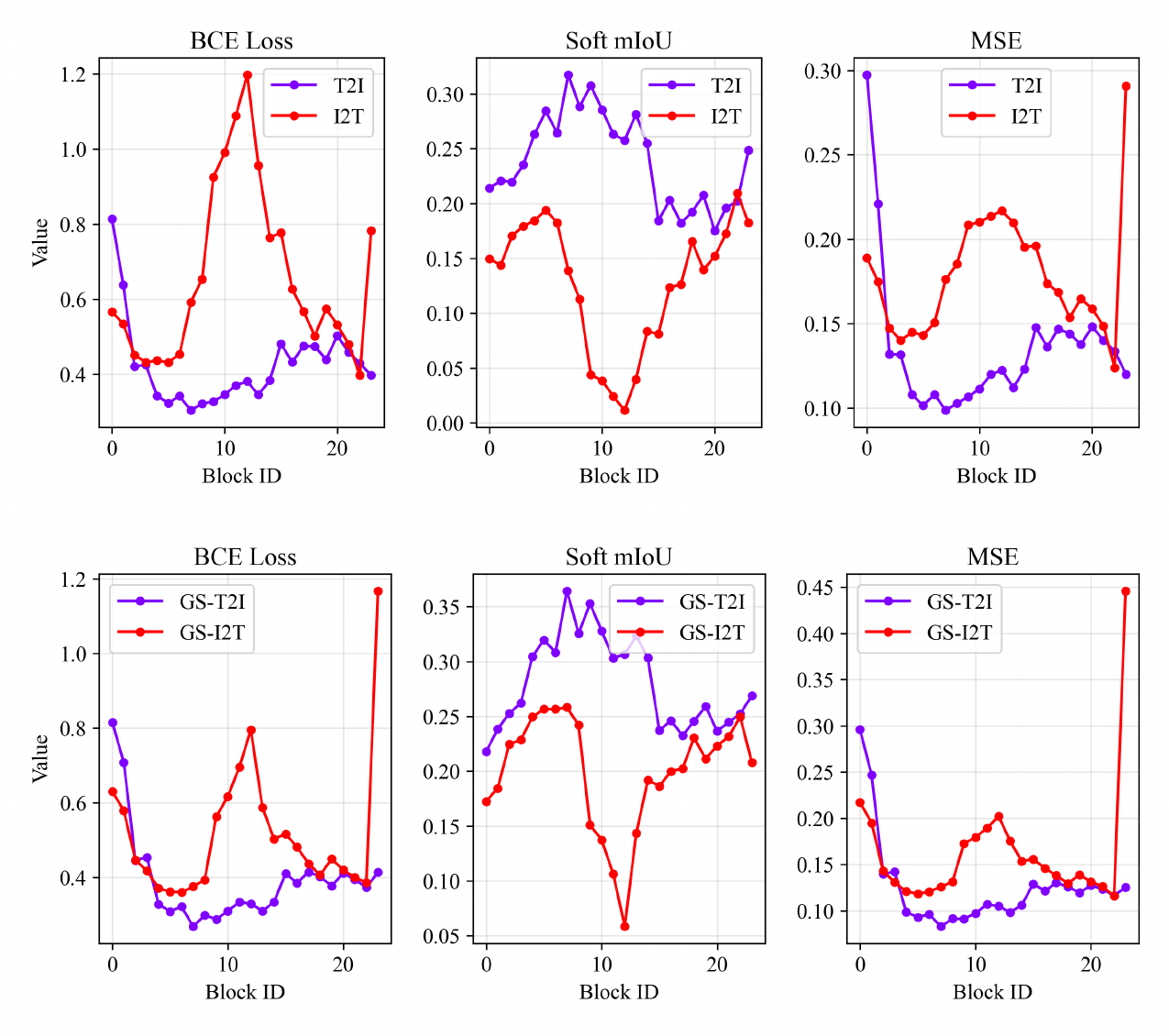
		}
		\caption{Transformer block analysis of SD3.5-M using Binary Cross Entropy Loss,
		Soft mIoU, and MSE, with Grounded SAM2 predictions as ground truth. Scores are
		shown without (upper) and with (lower) Gaussian smoothing.}
		\label{Fig:sd3.5-M-block-inspection}
	\end{figure}
	\begin{figure}
		\centering
		\includegraphics[width=\linewidth]{
			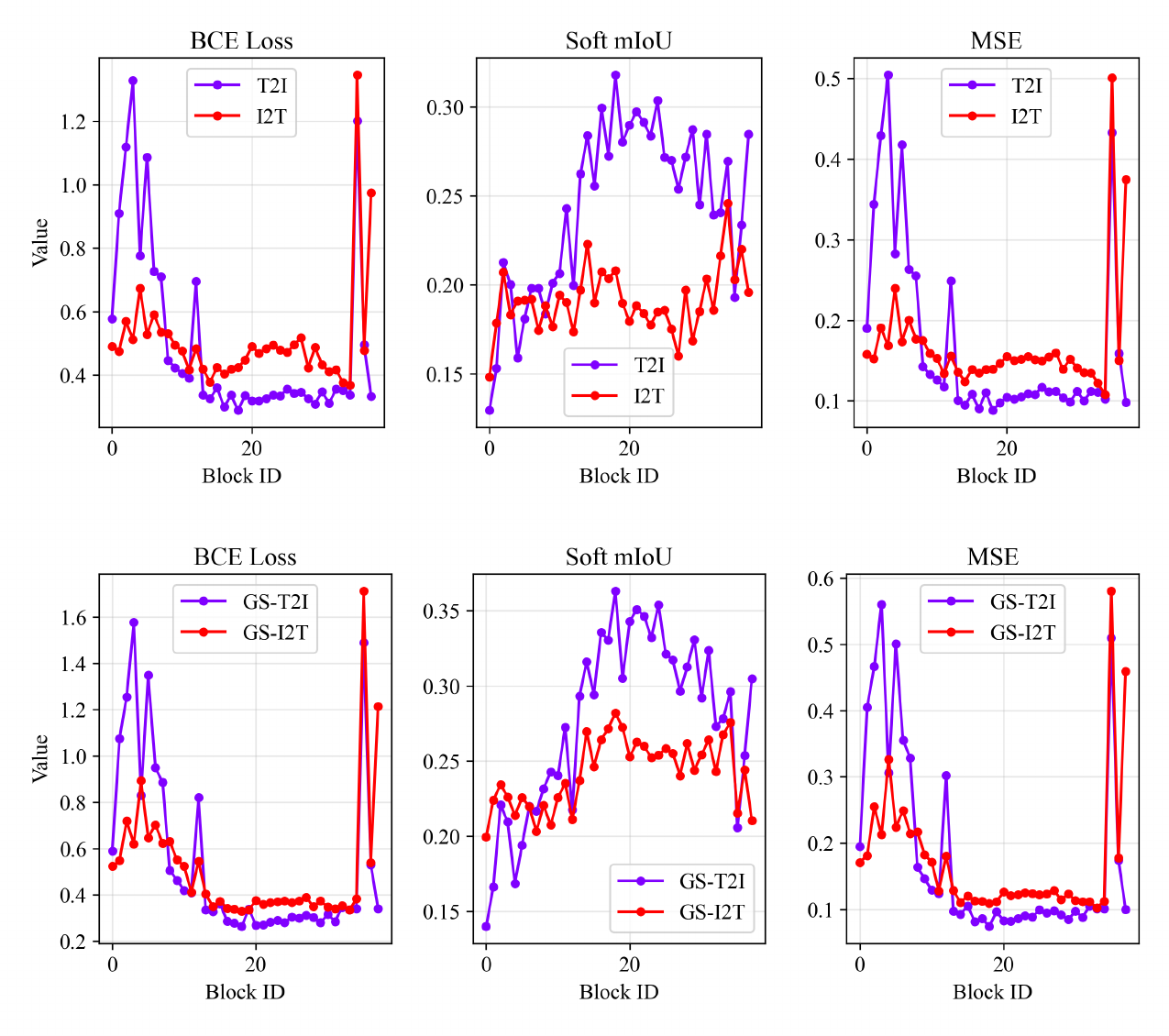
		}
		\caption{Transformer block analysis of SD3.5-L using Binary Cross Entropy Loss,
		Soft mIoU, and MSE, with Grounded SAM2 predictions as ground truth. Scores are
		shown without (upper) and with (lower) Gaussian smoothing.}
		\label{Fig:sd3.5-L-block-inspection}
	\end{figure}
	\begin{figure*}
		\centering
		\includegraphics[width=1\linewidth]{
			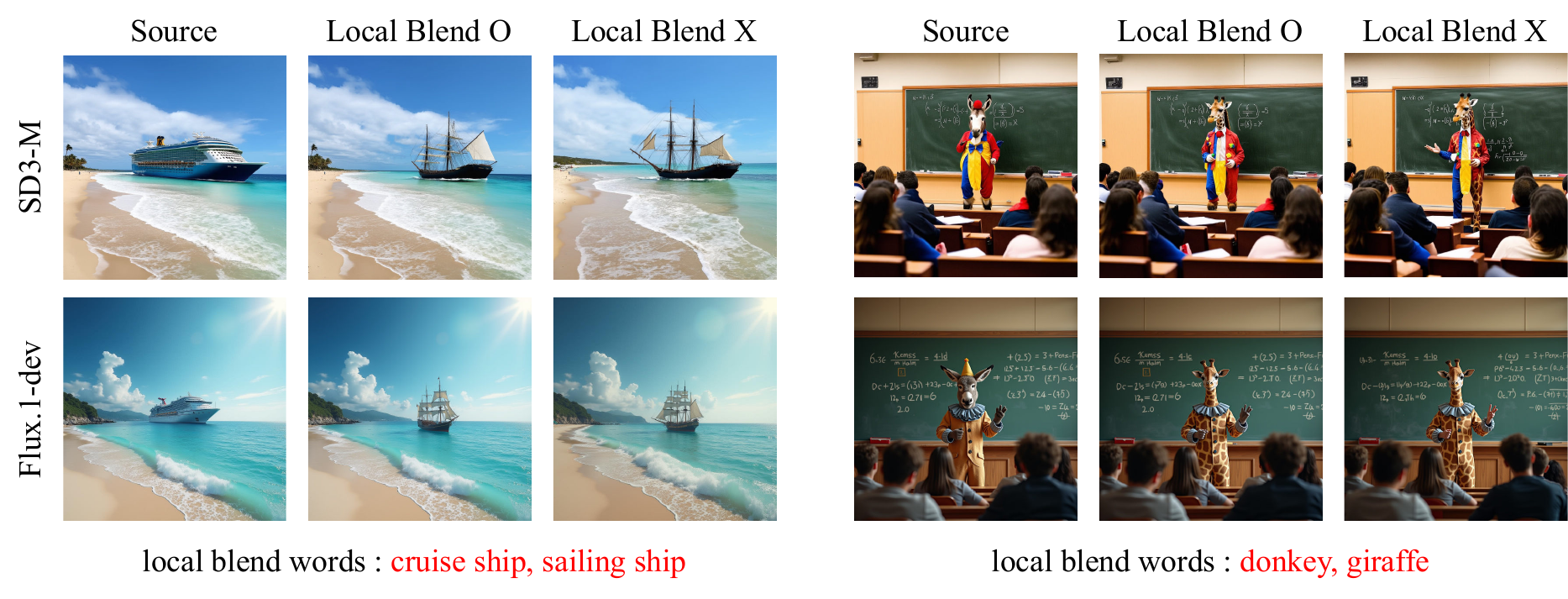
		}
		\caption{Local blending effects in SD3-M and Flux.1-dev models. The method
		excels at preserving non-targeted elements: in the maritime scene, the
		shoreline, island formations, and cloud patterns remain unchanged; in the classroom
		scene, the blackboard content and student arrangements are preserved. Here,
		we used the previously identified top-5 blocks from each model to generate
		masks with a local blending threshold of 0.4, applying the blending up to 50\%
		of total timestep iterations.}
		\label{Fig:sup_local_blend}
		\vspace{3mm}
	\end{figure*}

	\subsection{Identifying effective transformer blocks for obtaining clearer
	attention maps}
	\label{sec:sup_effective_blocks} In~\cref{Sec:attention-analysis-handling}, we
	presented our analysis of transformer blocks in Flux.1 to identify those
	producing clear attention maps suitable for local blending. Here, we extend
	this analysis to additional model architectures: SD3-M (\cref{Fig:sd3-M-block-inspection}),
	SD3.5-M (\cref{Fig:sd3.5-M-block-inspection}), and SD3.5-L (\cref{Fig:sd3.5-L-block-inspection}).
	We evaluate transformer blocks for each architecture using three metrics -
	Binary Cross Entropy Loss, Soft mIoU, and MSE - both with and without Gaussian
	smoothing. The top-5 blocks selected based on these metrics are summarized in~\cref{Tab:sup_smoothing_comparison}.

	As mentioned in the main paper, smaller models (SD3-M, SD3.5-M) largely
	maintain their block rankings regardless of Gaussian smoothing application. In
	contrast, larger models (SD3.5-L and Flux.1-dev) show significant changes in
	block rankings after smoothing. This suggests that while some blocks in larger
	models appear noisy in their raw form, they contain valuable structural
	information that becomes apparent after smoothing.

	\begin{figure*}
		\centering
		\includegraphics[width=1\linewidth]{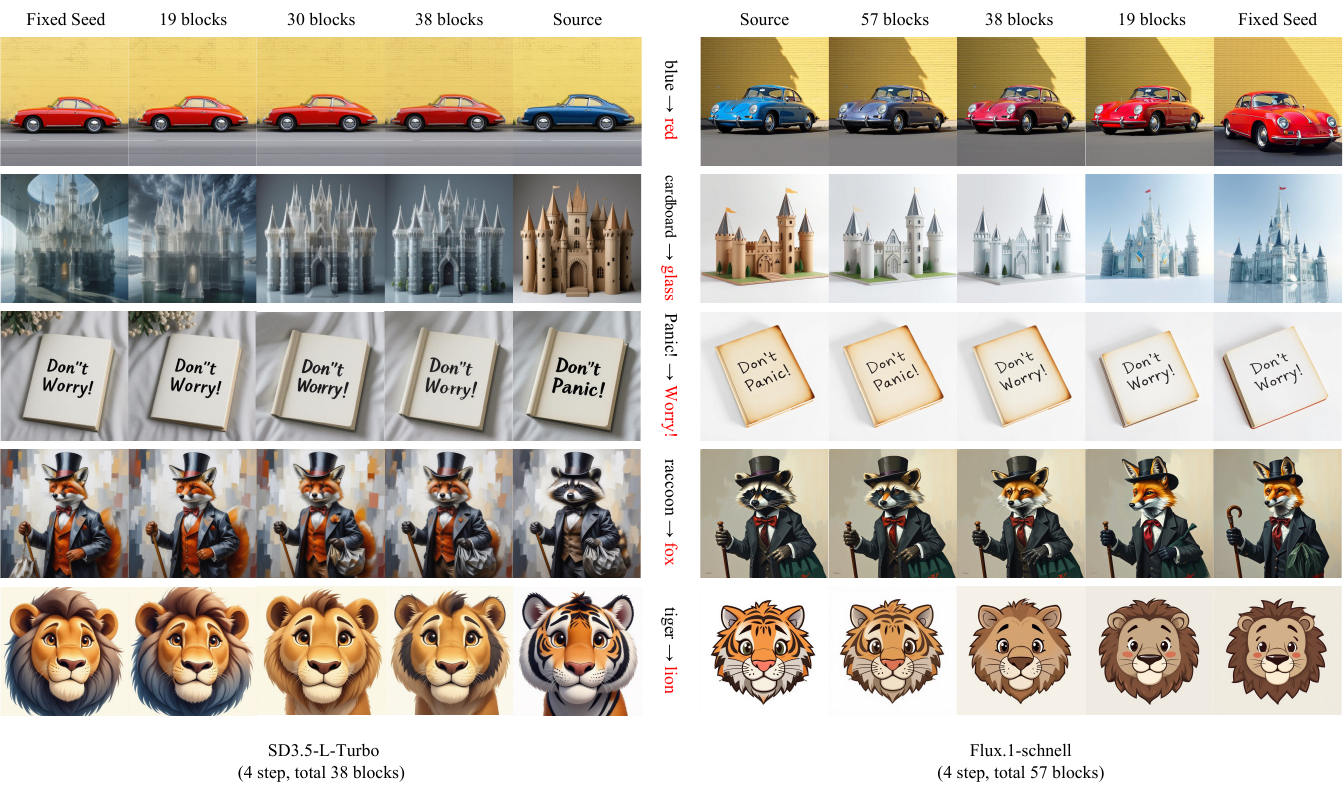}
		\caption{Qualitative results of SD3.5-L-Turbo (left) and Flux.1-schnell (right)
		demonstrating the impact of replacing block count (column) on edit strength,
		showing progression from source image through different block counts to fixed
		seed (corresponding to block count 0). Decreasing the number of replaced
		blocks strengthens the edit effect while reducing structural similarity and
		style to the source image. Local blending was not used to better focus on the
		impact of block replacement.}
		\label{Fig:sup_block_impact}
		\vspace{3mm}
	\end{figure*}
	We utilize the T2I portions of the identified top-5 transformer blocks for
	local blending operations. We selectively compute full attention maps only for
	these 5 blocks while using PyTorch's optimized SDPA kernel for all other blocks.
	This approach achieves both precise attention control and computational
	efficiency. The resulting attention maps from these selected blocks serve as the
	foundation for our local blending mechanism, enabling precise and controlled
	image editing. Qualitative results with and without local blending are shown in
	\cref{Fig:sup_local_blend} for comparison.

	\subsection{Impact of block selections on edit strength}
	\label{sec:sup_additional_ablations} As discussed in~\cref{Sec:editing-block-selection},
	the number of replacing blocks can serve as a hyperparameter to control edit
	strength. \cref{Fig:sup_block_impact} presents ablation studies on two MM-DiT few-step
	models, where we varied the number of replacing blocks from the initial block.
	A notable observation is that some generated images exhibit excessive similarity
	to the source image, even with one timestep injection of our method, which constrains
	the model's editing capabilities. Given that further reduction in timesteps is
	impossible, we investigated adjusting the timestep scheduler to mitigate this similarity;
	however, this approach also proved ineffective in addressing the limitations.
	In this context, block control emerges as a particularly effective solution for
	4-step distilled models, SD3.5-L-Turbo and Flux.1-schnell, with the latter showing
	a more pronounced effect. Through empirical investigation, we find that replacing
	blocks 38 and 30 yields favorable results for Flux.1-schnell and SD3.5-L-Turbo,
	respectively.

	\section{User Study and Additional Qualitative Results}

	To address the limitations of LPIPS and CLIP scores in capturing nuanced edit quality,
	we conducted a comprehensive user study using samples from~\cref{Tab:quantitative_comparison}.
	The study evaluated three widely-used models (SD3-M, Flux.1-dev, and Flux.1-schnell)
	with at least 30 participants per model (96 participants total). For fair
	comparison, all results were generated using purely $\mathbf{q}_{i}$,
	$\mathbf{k}_{i}$ replacement without manual per-sample local blending, though it
	would further enhance outcomes. Our findings reveal that our method uniquely
	balances strong target alignment (akin to direct generation, which sacrifices
	preservation contrastingly) with content preservation (comparable to prompt-change,
	which however fails to implement the edit).

	\begin{figure}
		\centering
		\includegraphics[width=\linewidth]{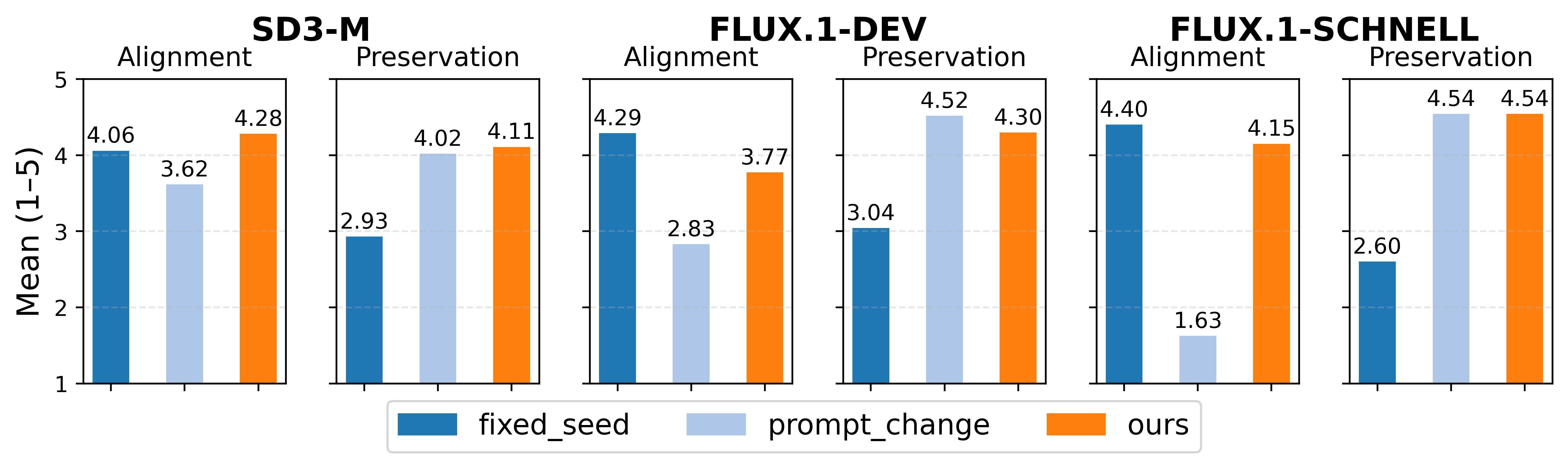}
		\caption{User study results comparing editing quality across different
		methods. Participants evaluated images based on target prompt alignment and source
		content preservation. Our method demonstrates superior balance between
		achieving desired edits while maintaining original image characteristics, outperforming
		both fixed seed generation (high prompt alignment, poor identity preservation)
		and prompt switching (good identity preservation, weak editing effects).}
		\label{Fig:user_study_results}
	\end{figure}

	Beyond the user study validation, we present extensive qualitative results (\cref{Fig:sup_additional_qual1},
	\cref{Fig:sup_additional_qual2}) from our benchmark experiments discussed in~\cref{Sec:experiments},~\cref{Tab:quantitative_comparison},
	and~\cref{Fig:user_study_results}. These results showcase the robustness of
	our core approach across diverse editing scenarios using only input projection
	replacement, without hyperparameter tuning for individual cases. Additionally,
	\cref{Fig:sup_aesthetic} demonstrates exemplary cases where local blending was
	applied to achieve enhanced visual quality and editing precision.

	\begin{figure}
		\centering
		\includegraphics[width=\linewidth]{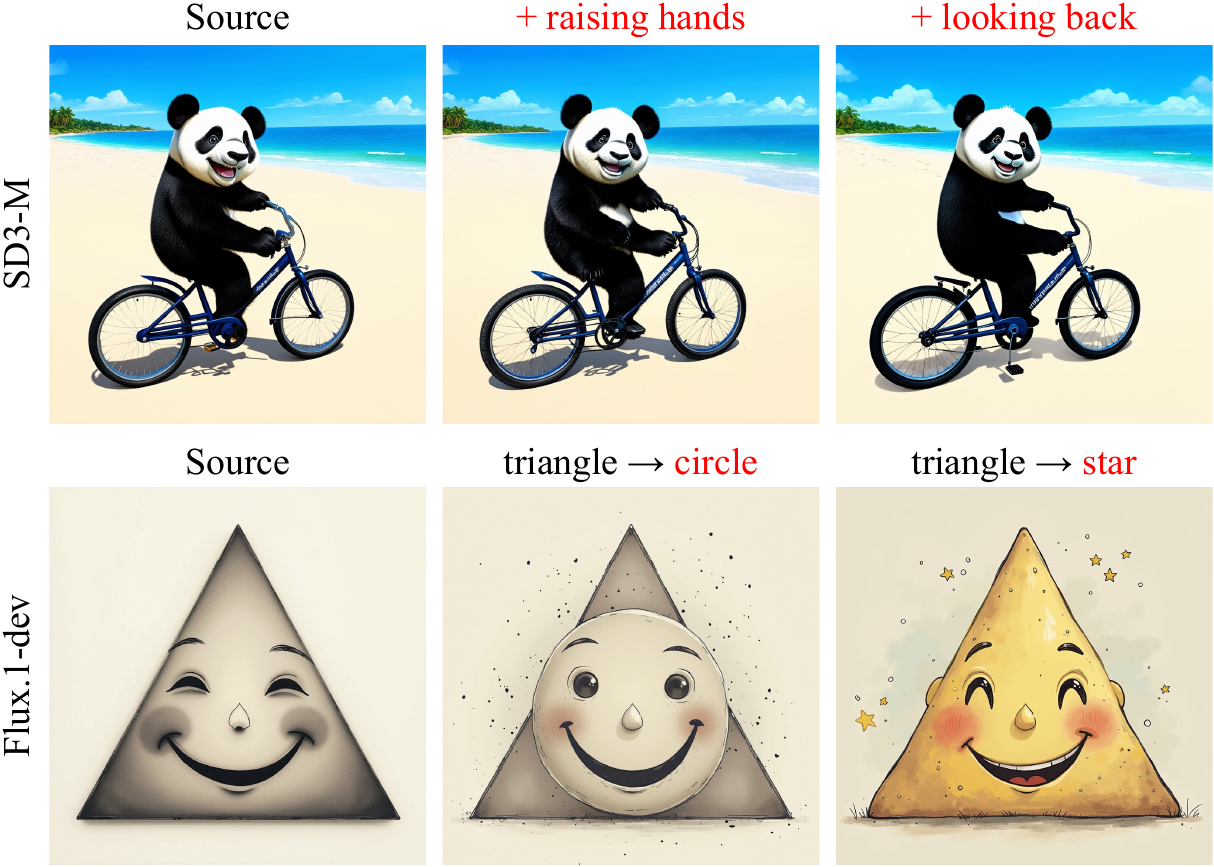}
		\caption{Limitations of our method. As noted in~\cref{Sec:conclusion}, our
		method is unable to achieve identity-preserving non-rigid transformations
		like those demonstrated in MasaCtrl~\cite{cao2023masactrl}. Despite exploring
		various strategies to effectively modify self-attention related components,
		we found it challenging to develop a robust method that could replicate the
		capabilities of prior works using U-Net backbones. Since our method operates
		by replacing target input projections with source input projections during early
		timesteps, modifying low-level structures that are determined in these
		stages can occasionally be challenging, particularly in few cases involving colors
		and rough geometric layouts. While manually adjusting replacing timesteps
		can mitigate this issue, we leave the development of a more systematic solution
		for future work.}
		\label{Fig:sup_limitation}
	\end{figure}

	\section{Comparison with Other Recent Works}
	To contextualize our contribution, this section discusses several recent and concurrent
	research efforts that have emerged alongside the effectiveness of modern MM-DiT
	architectures and RF formulations. These works often aim to improve real-image
	editing by building upon more accurate inversion methods and manipulating
	internal features. For instance, RF-Solver~\cite{wang2024taming} uses a Taylor
	expansion to derive a more precise ODE solution and reduce inversion errors. Similarly,
	FireFlow~\cite{deng2024fireflow} proposes an efficient few-step numerical
	solver that achieves second-order accuracy at first-order computational cost
	by reusing intermediate velocities. Both of these works also suggest swapping value
	features from the source branch into the target branch's self-attention layers
	to better preserve original content. Other approaches like FluxSpace~\cite{dalva2024fluxspace}
	define a semantic representation space from the projected value features post-attention,
	and interpolate within this space to add controls. Additionally, StableFlow~\cite{avrahami2025stable}
	identifies that not all layers in MM-DiT contribute equally to image formation
	and proposes a method to find a sparse set of ``vital layers'' crucial for the output.
	This concept, while not a direct equivalent, resonates with our finding of
	selecting optimal blocks with less noisy attention maps.

	In contrast to these approaches, our paper's fundamental contribution is an architectural
	analysis of MM-DiT's attention mechanisms. We explore how principles from
	prior models like U-Net can be effectively transferred, leading to a precise,
	prompt-based editing method tailored to modern MM-DiT architectures. This attention-centric
	analysis provides unique insights orthogonal to the aforementioned methods,
	enabling detailed edits through attention control that function even without
	inversion, yet yielding superior results when paired with techniques like RF inversion
	as shown in our experiments.

	\section{Limitations and Future Directions}
	\label{sec:sup_limitation} Our approach enables precise attention control through
	optimal transformer block selection and targeted input projection modifications.
	However, two limitations persist: the need for empirical parameter tuning in
	local blending and the inability to support identity-preserving non-rigid
	transformations (\cref{Fig:sup_limitation}). Beyond these technical limitations,
	we observe promising potential in applying these models to visual grounding
	and segmentation tasks, given their ability to capture abstract attributes transcending
	conventional object boundaries. We leave these challenges as potential
	directions for future research.

	\section{Used Prompts}
	In this section, we provide the list of prompts used to generate the main paper
	figures, where they were not explicitly stated in the text. Relevant codes for
	reproducing our results will be open-sourced upon publication. Due to space
	constraints, additional prompts used for benchmarking and supplemental figures
	will be available in our public repository.

	\noindent
	\textbf{Figure 1.}
	\begin{itemize}
		\item Source 1: \textit{``beautiful oil painting of a steamboat in a river
			in the afternoon. On the side of the river is a large brick building with a
			sign on top that says `SD3'''}

		\item Target 1: \textit{``beautiful oil painting of a steamboat in a river
			in the afternoon. On the side of the river is a large brick building with a
			sign on top that says `FLUX'''}

		\item Source 2: \textit{``Detailed pen and ink drawing of a happy giraffe
			butcher selling meat in its shop''}

		\item Target 2: \textit{``Detailed pen and ink drawing of a happy dragon
			butcher selling meat in its shop''}

		\item Source 3: \textit{``A photograph of the inside of a subway train. There
			are frogs sitting on the seats. One of them is reading a newspaper. The window
			shows the river in the background''}

		\item Target 3: \textit{``A photograph of the inside of a subway train. There
			are rabbits sitting on the seats. One of them is reading a newspaper. The window
			shows the river in the background''}

		\item Source 4: \textit{``a guy in the forest with a sword and shield, fighting
			a dragon, holding a large sign `help me'''}

		\item Target 4: \textit{``a guy in the forest with a sword and shield, fighting
			a dragon, holding a large sign `Please don't kill me'''}

		\item Source 5: \textit{``A crab made of cheese on a plate''}

		\item Target 5: \textit{``A cartoon-style drawing of a crab made of cheese
			on a plate''}

		\item Source 6: \textit{``translucent pig, inside is a smaller pig''}

		\item Target 6: \textit{``translucent whale, inside is a smaller whale''}

		\item Source 7: \textit{``A 4K DSLR image of a Hound dog dressed in a finely
			tailored houndstooth check suit with bold, oversized patterns standing on a
			perfectly manicured grassy field holding a beautifully crafted banner that
			says `Go Puppy Team!'\,''}

		\item Target 7: \textit{``A 4K DSLR image of a Zebra dressed in a finely
			tailored zebra-striped suit with bold, oversized patterns standing on a perfectly
			manicured grassy field holding a beautifully crafted banner that says `Go
			Zebra Team!'\,''}

		\item Source 8: \textit{``A mischievous ferret with a playful grin squeezes
			itself into a large glass jar, surrounded by colorful candy. The jar sits
			on a wooden table in a cozy kitchen, and warm sunlight filters through a nearby
			window''}

		\item Target 8: \textit{``A mischievous lion with a playful grin squeezes
			itself into a large glass jar, surrounded by colorful candy. The jar sits
			on a wooden table in a cozy kitchen, and warm sunlight filters through a nearby
			window''}
	\end{itemize}

	\noindent
	\textbf{Figure 3.}
	\begin{itemize}
		\item \textit{``a panda riding a bicycle on the beach under blue sky''}
	\end{itemize}

	\noindent
	\textbf{Figure 4.}
	\begin{itemize}
		\item \textit{``a photograph of a fiddle next to a basketball on a ping pong
			table''}
	\end{itemize}

	\noindent
	\textbf{Figure 6.}
	\begin{itemize}
		\item \textit{``a cute tiger driving a sports car under starry night with
			blue moon in new york''}
	\end{itemize}

	\noindent
	\textbf{Figure 8.}
	\begin{itemize}
		\item Source: \textit{``a drawing of a series of musical notes wrapped
			around the Earth''}

		\item Target: \textit{``a drawing of a series of musical notes wrapped
			around the Moon''}
	\end{itemize}

	\noindent
	\textbf{Figure 10.}
	\begin{itemize}
		\item Source: \textit{``a panda riding a bicycle on the beach under blue sky''}

		\item Target 1: \textit{``a princess with a crown riding an elephant on the
			beach under blue sky''}

		\item Target 2: \textit{``a squirrel with a baseball cap riding a blue
			motorbike on the beach under blue sky''}

		\item Target 3: \textit{``a cute hamburger with fried chicken legs riding a
			green motorbike in the grand canyon under blue sky''}
	\end{itemize}

	\noindent
	\textbf{Figure 11.}
	\begin{itemize}
		\item Source: \textit{``A whimsical scene featuring a playful hybrid
			creature: a hippopotamus with golden, crispy waffle-textured skin, lounging
			in a surreal habitat blending water and breakfast elements like giant utensils
			and plates.''}

		\item Target: \textit{``A whimsical scene featuring a playful hybrid
			creature: an elephant with golden, crispy waffle-textured skin, lounging in
			a surreal habitat blending water and breakfast elements like giant utensils
			and plates.''}
	\end{itemize}

	\noindent
	\textbf{Figure 12.}
	\begin{itemize}
		\item Source 1: \textit{``beautiful oil painting of a steamboat in a river
			in the afternoon. On the side of the river is a large brick building with a
			sign on top that says `SD3'''}

		\item Target 1: \textit{``beautiful oil painting of a steamboat in a river
			in the afternoon. On the side of the river is a large brick building with a
			sign on top that says `FLUX'''}

		\item Source 2: \textit{``a cat sitting on a stairway railing''}

		\item Target 2: \textit{``a squirrel sitting on a stairway railing''}
	\end{itemize}

	\noindent
	\textbf{Figure 15.}
	\begin{itemize}
		\item Source 1: \textit{``a grandmother reading a book to her grandson and
			granddaughter''}

		\item Target 1: \textit{``a grandmother reading a holographic storybook to
			her grandson and granddaughter in a floating space station''}

		\item Source 2: \textit{``three green peppers''}

		\item Target 2: \textit{``three red peppers''}

		\item Source 3: \textit{``A close-up high-contrast photo of Sydney Opera
			House sitting next to Eiffel tower, under a blue night sky of roiling energy,
			exploding yellow stars, and radiating swirls of blue''}

		\item Target 3: \textit{``A close-up high-contrast photo of Sydney Opera
			House sitting next to Eiffel tower, under a purple night sky of roiling energy,
			exploding yellow stars, and radiating swirls of purple''}

		\item Source 4: \textit{``a comic about two cats doing research''}

		\item Target 4: \textit{``a comic about two cats doing quantum physics
			research in a lab full of glowing experiments''}

		\item Source 5: \textit{``a cartoon of a bear birthday party''}

		\item Target 5: \textit{``a cartoon of a panda birthday party''}

		\item Source 6: \textit{``a cat patting a crystal ball with the number 7
			written on it in black marker''}

		\item Target 6: \textit{``a cat patting a crystal ball with the number 13
			written on it in black marker''}
	\end{itemize}

	\begin{figure*}
		\centering
		\includegraphics[width=1\linewidth]{
			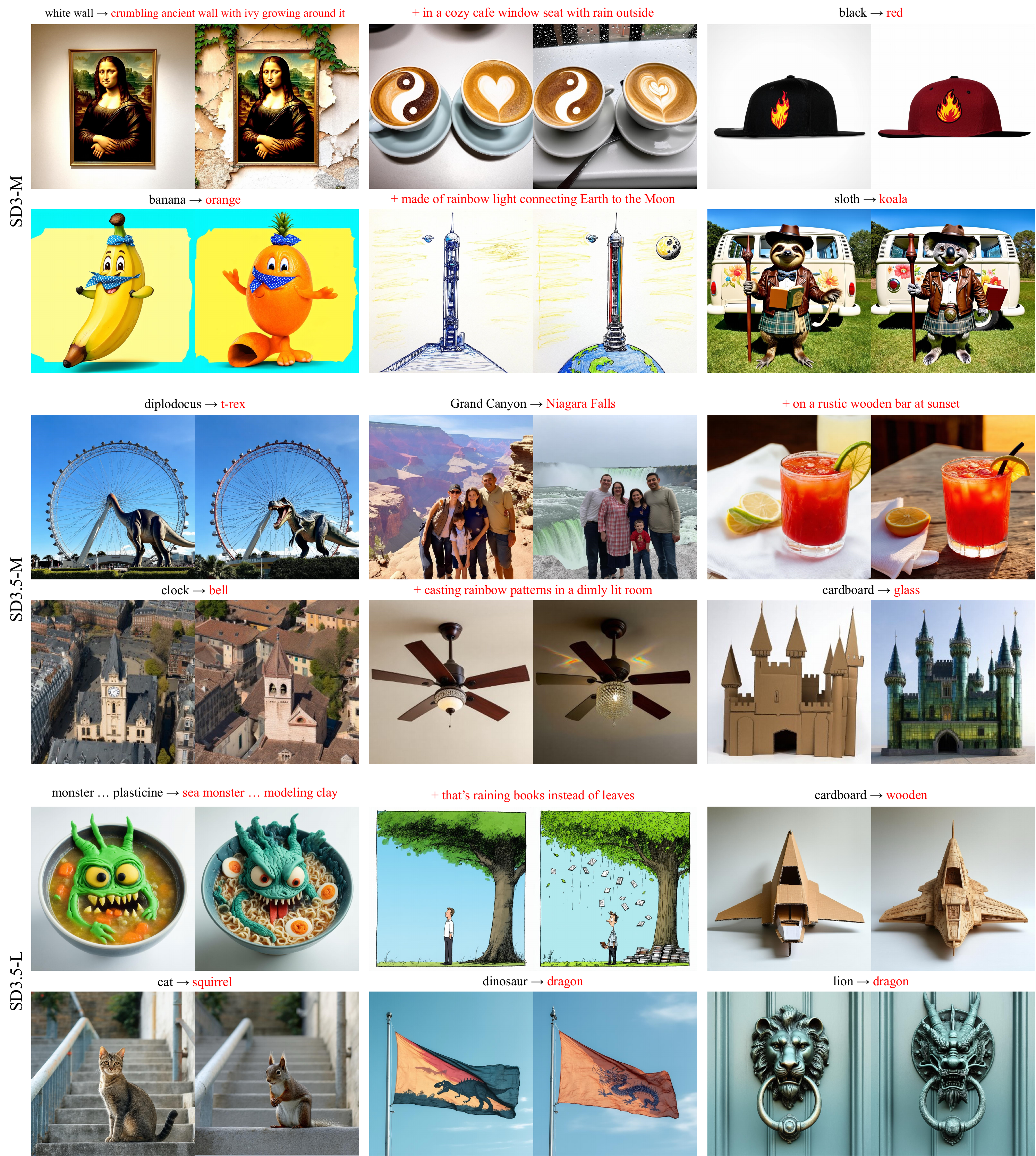
		}
		\caption{Additional qualitative results from experiments reported in~\cref{Tab:quantitative_comparison},
		demonstrating edits across diverse model variants. All results shown use
		only input projection replacements ($\mathbf{q}_{i}$, $\mathbf{k}_{i}$),
		without local blending operation.}
		\label{Fig:sup_additional_qual1}
	\end{figure*}
	\begin{figure*}
		\centering
		\includegraphics[width=1\linewidth]{
			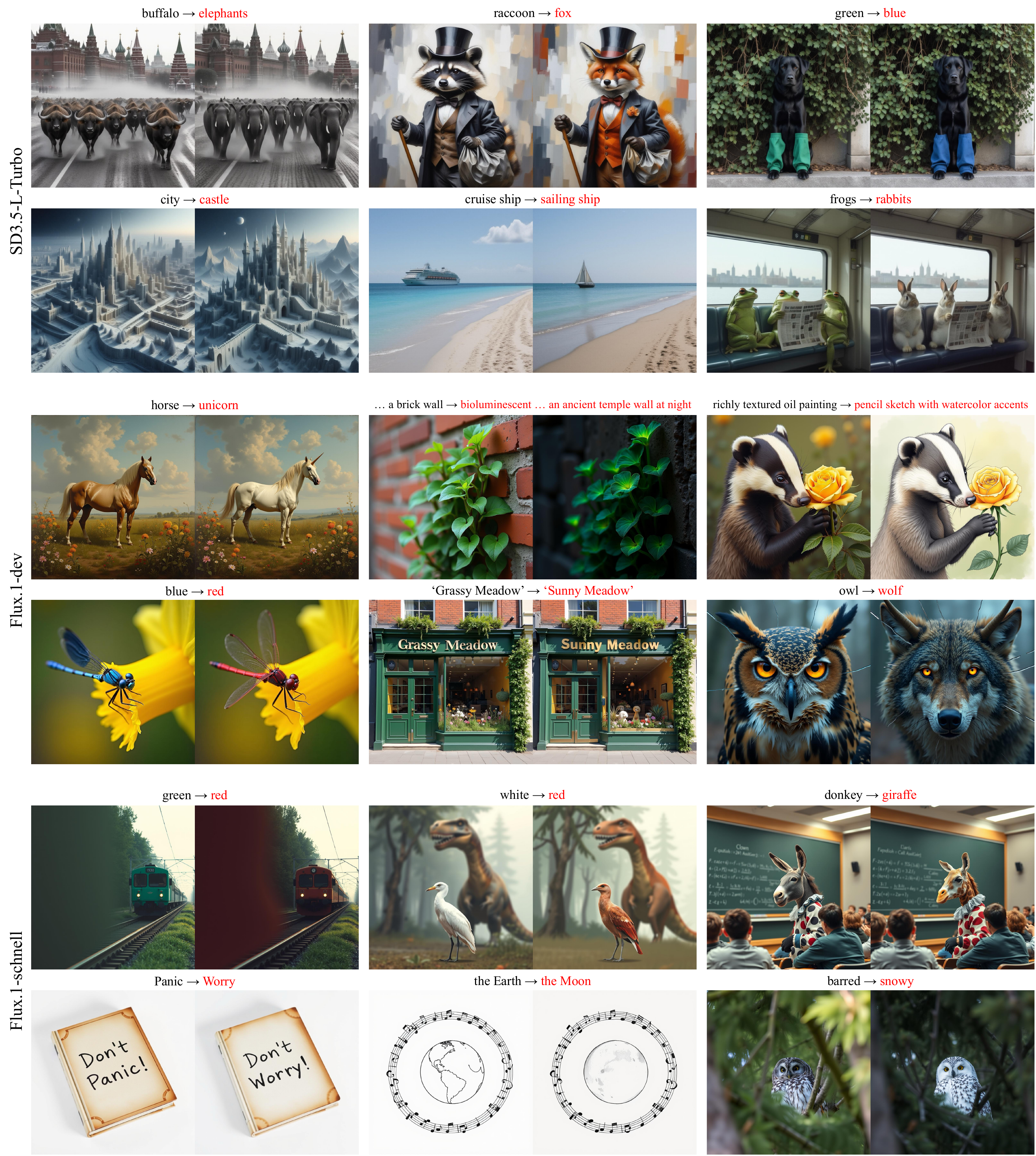
		}
		\caption{Additional qualitative results from experiments reported in~\cref{Tab:quantitative_comparison},
		demonstrating edits across diverse model variants. All results shown use
		only input projection replacements ($\mathbf{q}_{i}$, $\mathbf{k}_{i}$),
		without local blending operation.}
		\label{Fig:sup_additional_qual2}
	\end{figure*}
	\begin{figure*}
		\centering
		\includegraphics[width=1\linewidth]{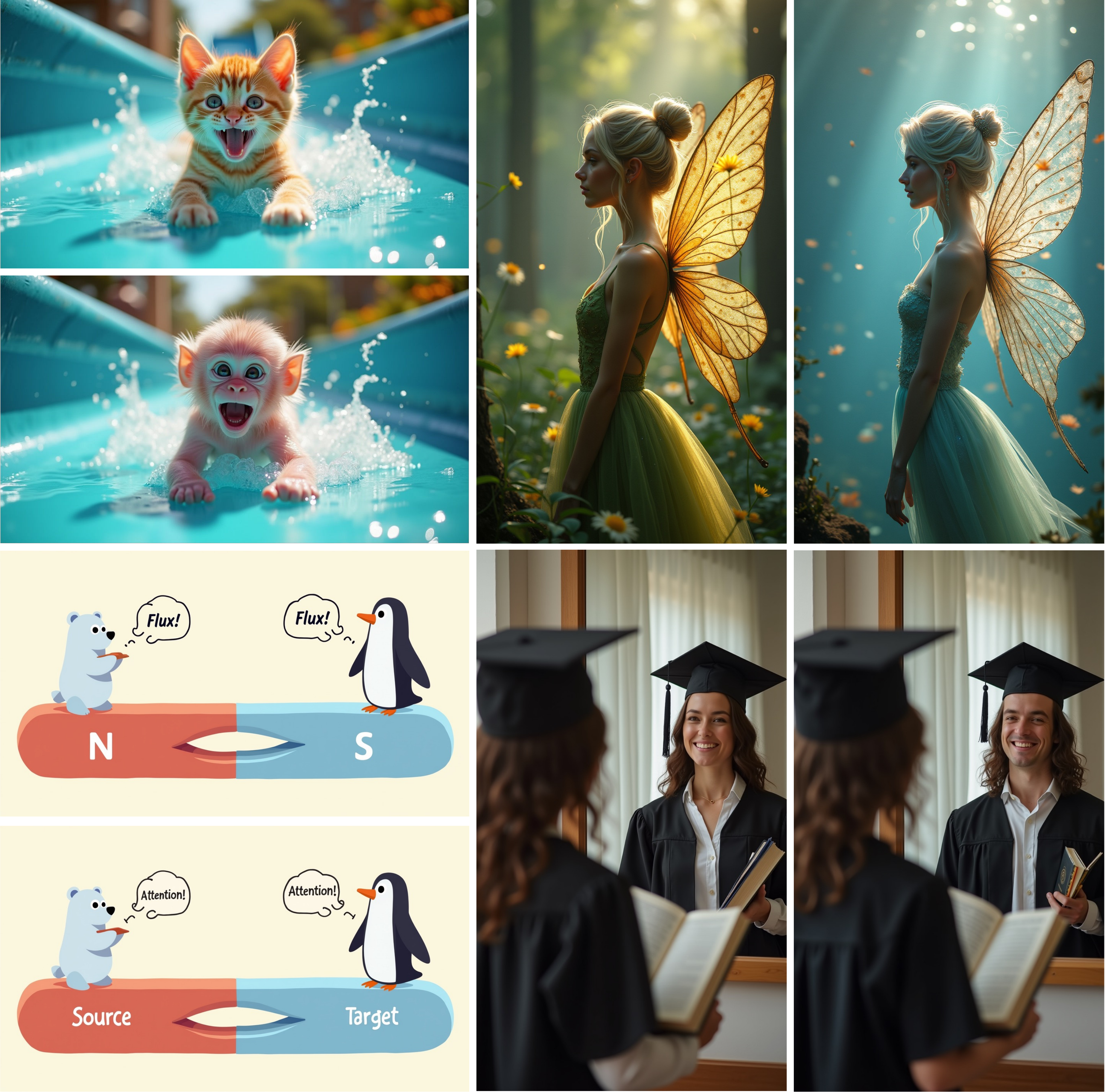}
		\caption{Additional qualitative results of our method, showcasing various editing
		scenarios: (a) changing `orange kitten' into `pink monkey', (b) converting `a
		forest fairy' into `an ocean fairy', (c) modifying text from `N, S, and Flux!'
		to `Source, Target, and Attention!', and (d) changing the identity of subjects,
		such as transforming a `grown woman' into a `grown man'.}
		\label{Fig:sup_aesthetic}
	\end{figure*}

	\begin{figure*}
		\centering
		\includegraphics[width=0.93\linewidth]{
			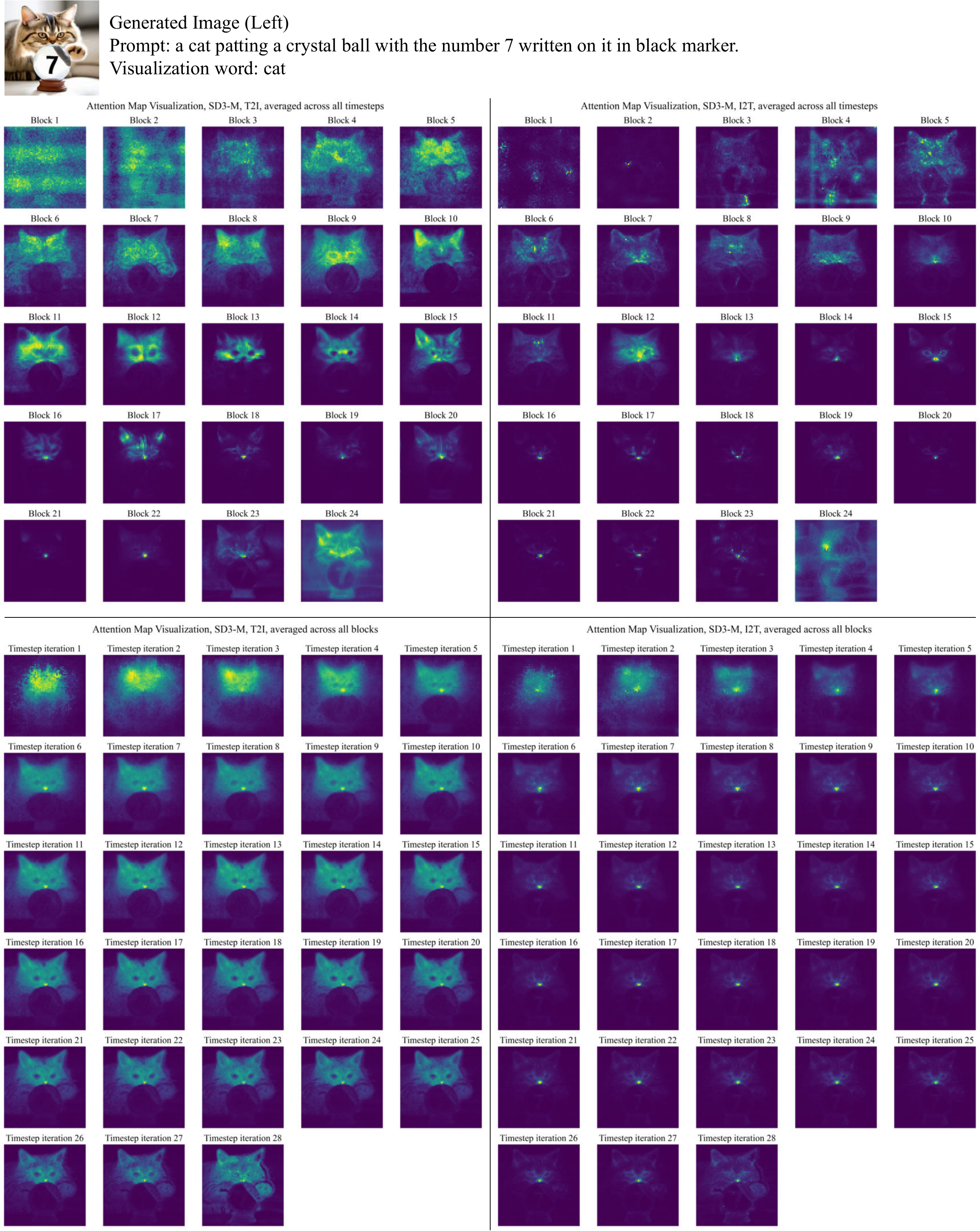
		}
		\caption{Visualization of T2I (left) and I2T (right) attention maps in SD3-M.
		Upper rows show per-block attention maps (averaged across 28 timesteps),
		while lower rows show per-timestep attention maps (averaged across all blocks).
		T2I portions generally capture semantic concepts more effectively, though
		certain blocks exhibit significant noise. Timestep-wise analysis reveals that
		image structure and layout are primarily established in early denoising steps,
		supporting our approach of attention map replacement during only the first
		20\% of timesteps to preserve original image characteristics. Best viewed
		zoomed in.}
		\label{Fig:sup_attention_maps_sd3}
	\end{figure*}
	\begin{figure*}
		\centering
		\includegraphics[width=0.96\linewidth]{
			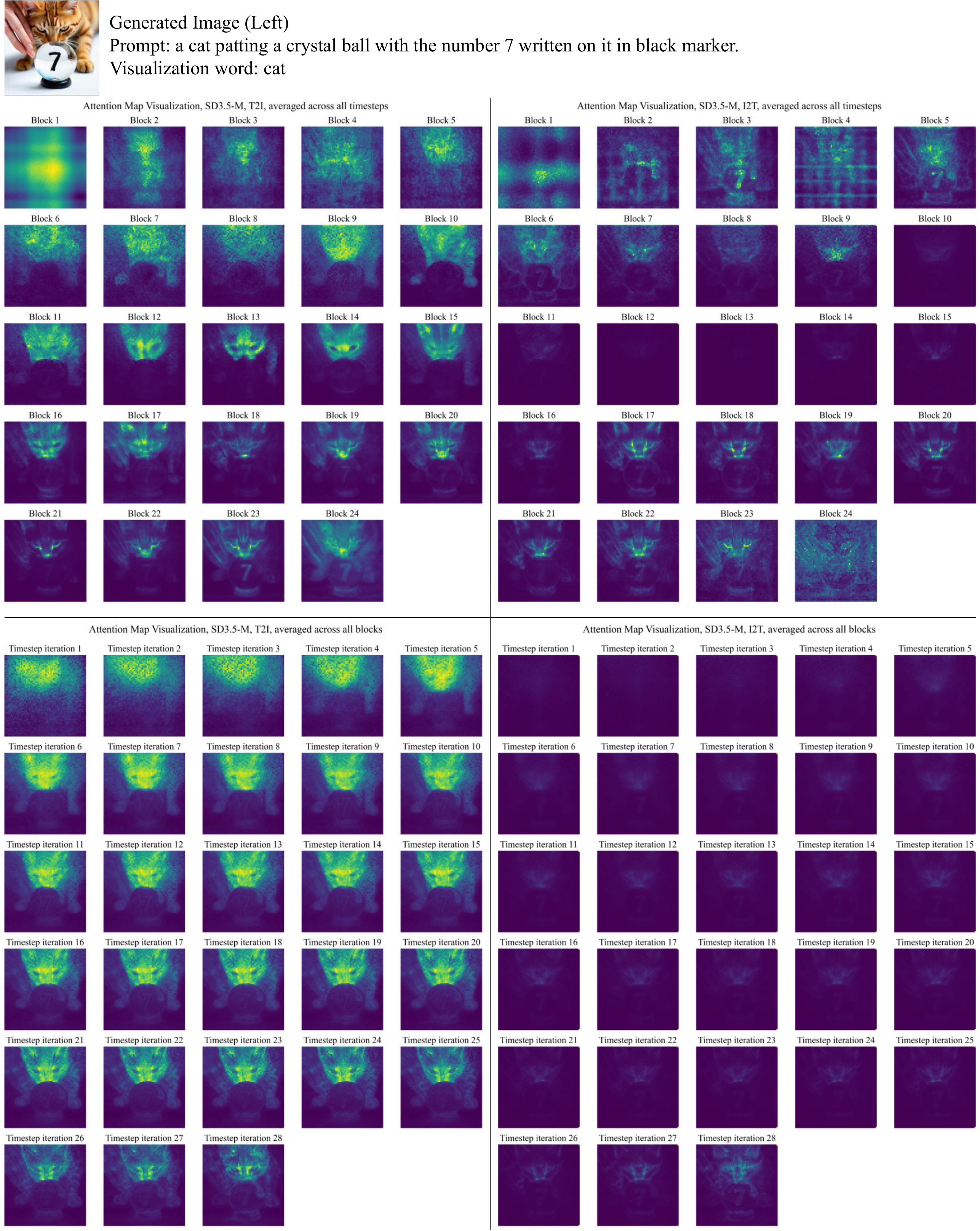
		}
		\caption{Visualization of T2I (left) and I2T (right) attention maps in SD3.5-M,
		averaged across timesteps (upper) and across transformer blocks (lower). The
		visualization format and observed patterns are mostly consistent with~\cref{Fig:sup_attention_maps_sd3}.
		Best viewed zoomed in.}
		\label{Fig:sup_attention_maps_sd3.5-M}
	\end{figure*}
	\begin{figure*}
		\centering
		\includegraphics[width=1\linewidth]{
			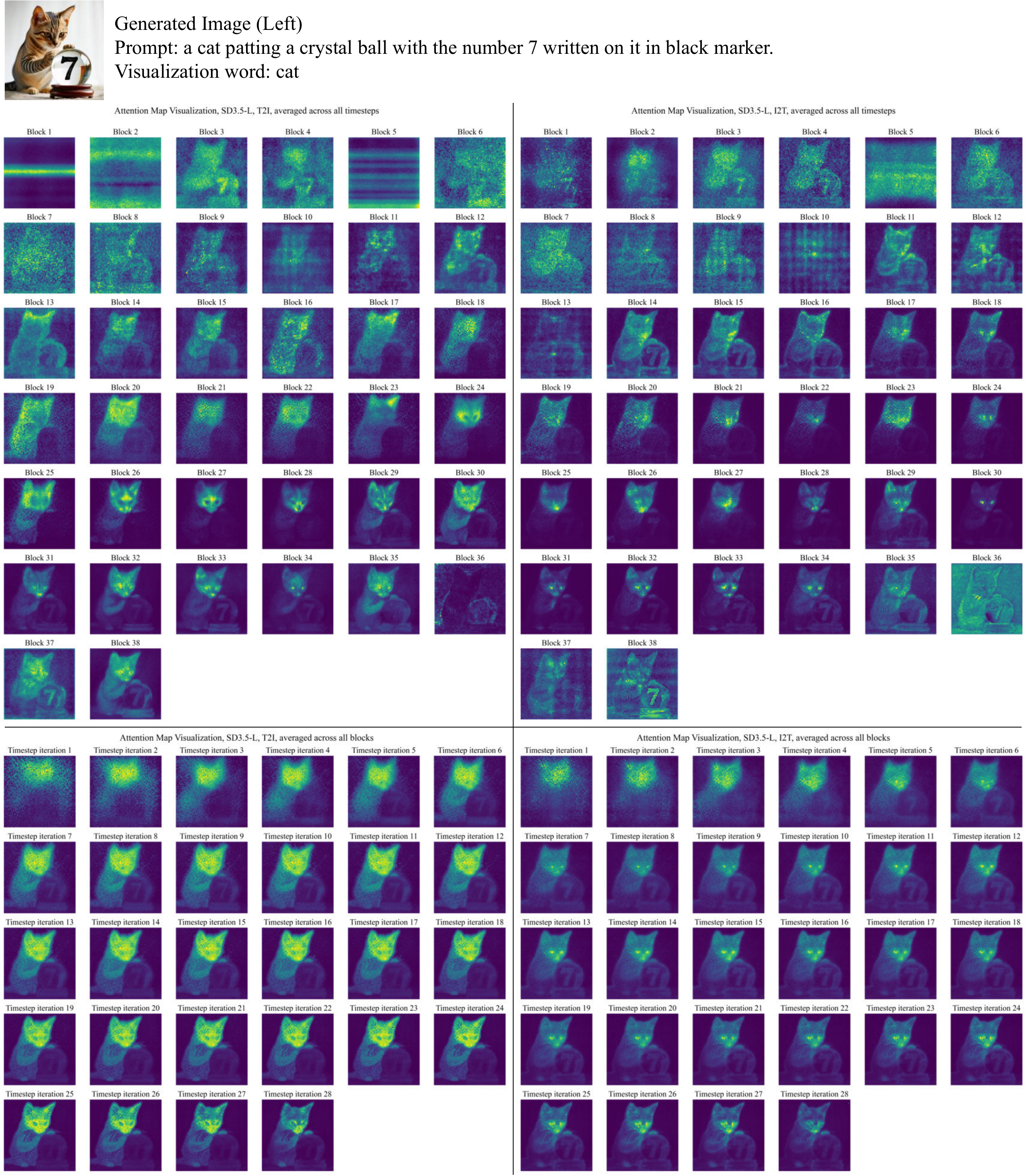
		}
		\caption{Visualization of T2I (left) and I2T (right) attention maps in SD3.5-L,
		averaged across timesteps (upper) and across transformer blocks (lower). The
		visualization format and observed patterns are mostly consistent with Fig.~\ref{Fig:sup_attention_maps_sd3},
		except that we observe noisier attention maps. Best viewed zoomed in.}
		\label{Fig:sup_attention_maps_sd3.5-L}
	\end{figure*}
	\begin{figure*}
		\centering
		\includegraphics[width=1.0\linewidth]{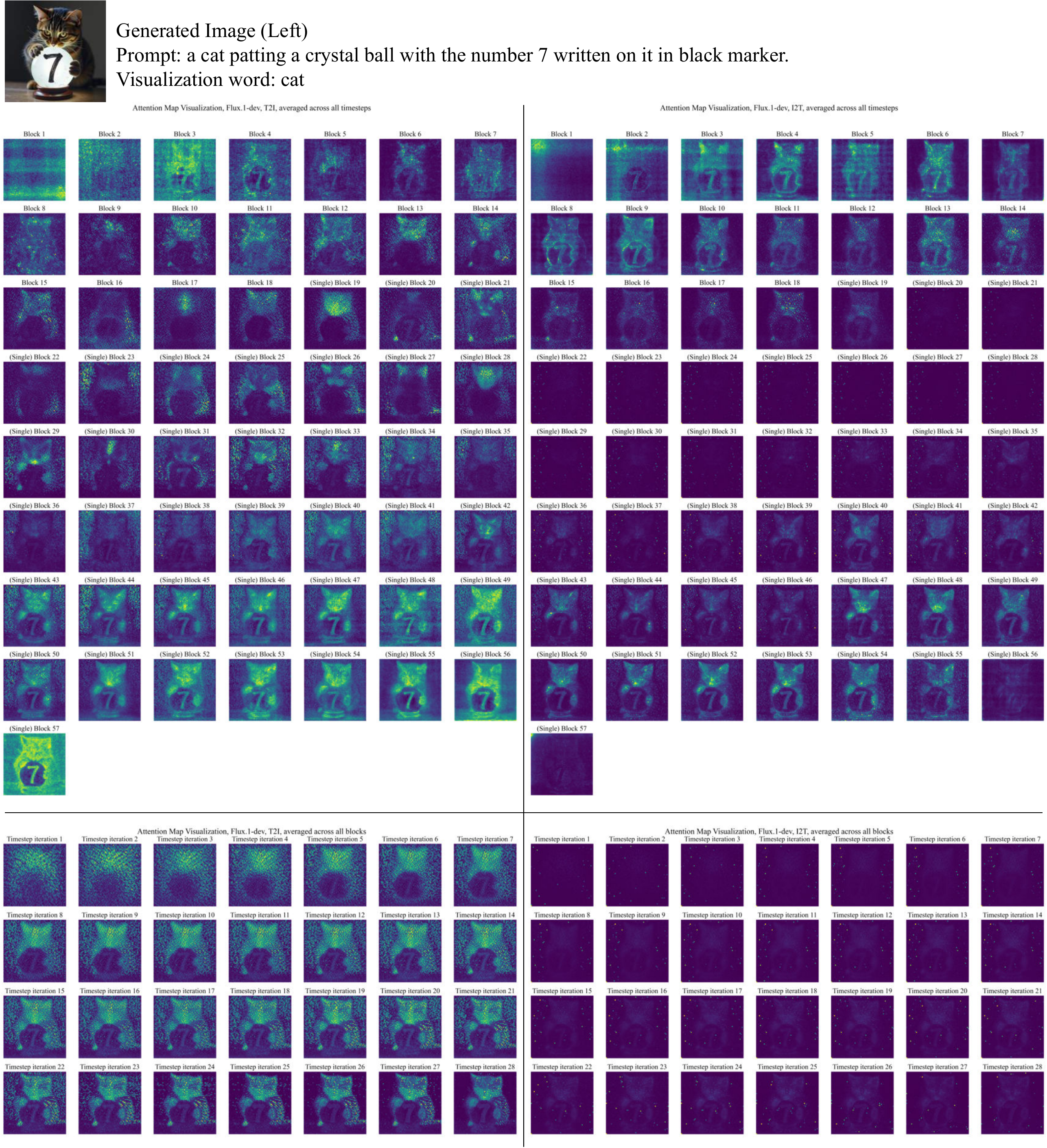}
		\caption{Visualization of T2I (left) and I2T (right) attention maps in Flux.1,
		averaged across timesteps (upper) and across transformer blocks (lower). The
		visualization format and observed patterns are mostly consistent with Fig.~\ref{Fig:sup_attention_maps_sd3}.
		It is worth noting that even in single-branch transformers, geometric and
		spatial patterns are preserved, indicating the preservation of information for
		each domain. Attention maps appear noisy in some blocks. Best viewed zoomed in.}
		\label{Fig:sup_attention_maps_flux}
		\vspace{3mm}
	\end{figure*}
	\begin{figure*}
		\centering
		\includegraphics[width=1\linewidth]{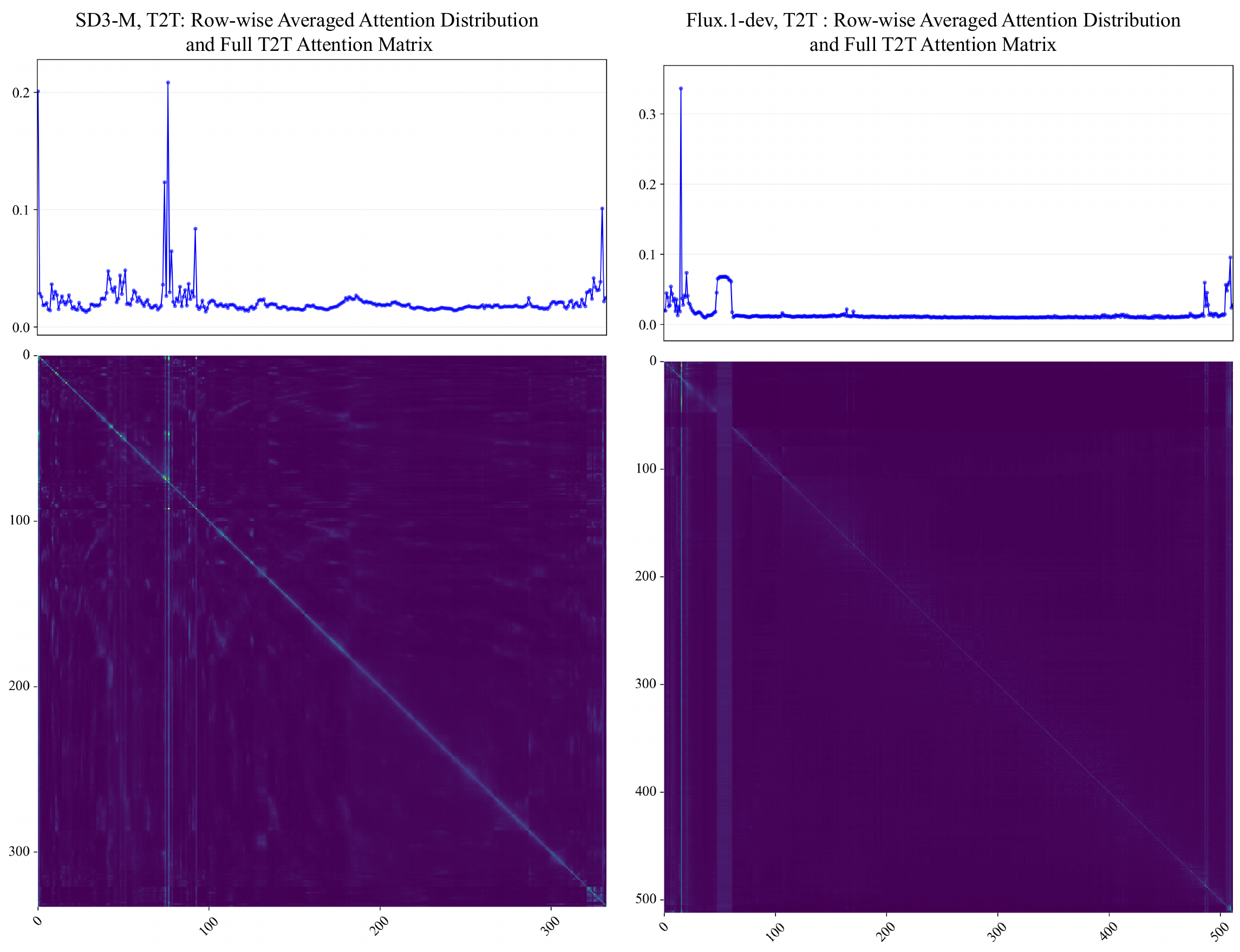}
		\caption{Visualization of the T2T portion of the attention maps in SD3-M (left)
		and Flux.1-dev (right). The heatmaps mostly show diagonal patterns, with stronger
		signals from special tokens. Above each heatmap, we present row-wise averaged
		attention values as line plots to better highlight the relative values among
		column indices.}
		\label{Fig:sup_t2t}
	\end{figure*}
\end{document}